\RequirePackage{xcolor}
\documentclass{iopjournal}
\usepackage{float}
\usepackage{TLC}
\usepackage{xurl}
\usepackage{booktabs}
\usepackage{url}
\usepackage[normalem]{ulem}
\usepackage{setspace}

\usepackage{enumitem}
\usepackage{subcaption}

\usepackage{mathtools}
\usepackage{gensymb}
\usepackage{amsmath}

\usepackage{graphicx, array, booktabs}
\usepackage{tabularx}   
\usepackage{subcaption} 
\newcolumntype{Y}{>{\centering\arraybackslash}p{0.40cm}} 

\usepackage[
  backend=biber,
  style=numeric-comp,
  sorting=none,
  maxbibnames=10,
  giveninits=true,
  url=false,
  date=year
]{biblatex}

\theoremstyle{remark}

\begin{document}
\articletype{paper}

\title{A Neuromorphic Architecture for  Scalable  Event-Based Control}

\author{Yongkang Huo$^1$$^4$, Fulvio Forni$^1$, Rodolphe Sepulchre$^1$$^2$$^3$}

\affil{$^1$The authors are with the University of Cambridge, Department of Engineering, Trumpington Street, CB2 1PZ.}

\affil{$^2$ R. Sepulchre is also with KU Leuven,
Department of Electrical Engineering (STADIUS),
KasteelPark Arenberg, 10,
B-3001 Leuven, Belgium,
\texttt{rodolphe.sepulchre@kuleuven.be}.}

\affil{$^3$The research leading to these results has received funding from the European Research Council under the Advanced ERC Grant Agreement SpikyControl n.101054323.}

\affil{$^4$ The work of Y. Huo was supported by the UK Engineering and Physical Sciences Research Council (EPSRC) grant 10671447 for the University of Cambridge Centre for Doctoral Training, the Department of Engineering.}

\email{\texttt{yh415@cam.ac.uk}, \texttt{f.forni@eng.cam.ac.uk}, \texttt{rs771@cam.ac.uk}}
\begin{abstract}

This paper introduces the ``rebound Winner-Take-All (RWTA)" motif as the basic element of a scalable neuromorphic control architecture. From the cellular level to the system level, the resulting architecture combines the reliability of discrete computation and the tunability of continuous regulation: it inherits the discrete computation capabilities of winner-take-all state machines and the continuous tuning capabilities of excitable biophysical circuits. The proposed event-based framework addresses continuous rhythmic generation and discrete decision-making in a unified physical modelling language. We illustrate the versatility, robustness,  and modularity of the architecture through the nervous system design of a snake robot. 
\end{abstract}

\section{Introduction}

The divide between the discrete and the continuous pervades today's control design and technology: ``high-level" system design is separated from ``low-level" control design, automation is separated from regulation, decision-making is separated from physical interaction. Discrete  methodologies use the discrete  language of finite state machines while continuous methodologies use the continuous modelling language of differential equations. The resulting need  to interface discrete automata and continuous differential equations has made the theory of hybrid  and cyberphysical systems a key component of control design. The advantages of separating the discrete and the continuous design are many. Yet, this separation has become a bottleneck in the design of scalable architectures that can control and regulate across a range of spatial and temporal scales \cite{sepulchre_control_2019,matni_quantitative_2024}. 

The divide between automation and regulation does not exist in animal machines. Nervous systems make decisions and regulate muscle activation with one and the same architecture that is event-based rather than continuous or discrete. Events occur in physical circuits obeying continuous-time differential equations, yet they can be counted and they can be modelled as the discrete events of a finite state machine  \cite{sepulchre_spiking_2022,ribar_neuromorphic_2021,schmetterling_neuromorphic_2024}. Neuromorphic control \cite{ribar_neuromorphic_2021} aims at mixing discrete computation and continuous regulation by borrowing  the physical principles and the architecture of excitable neuronal circuits. These neuromorphic approaches build upon four decades of progress in analogue and mixed-signal VLSI that have turned Carver Mead’s vision of “computing with physics’’ into wafer-scale neuromorphic substrates capable of emulating  large-scale neural networks
\cite{mead_analog_1989,indiveri_winner-take-all_1997,indiveri_neuromorphic_2011,
benjamin_neurogrid_2014,furber_spinnaker_2014,kudithipudi_neuromorphic_2025}.  

Earlier work in neuromorphic control by the authors has focused on the control and modulation of  single neurons \cite{ribar_neuromorphic_2021} or small circuits of  few neurons \cite{schmetterling_neuromorphic_2024}. In contrast, the aim of the present paper is to introduce a {\it scalable} neuromorphic control architecture for the design of hierarchical event-based machines. 

The core element of our methodology is to combine the ``rebound excitability" of isolated biophysical neurons with the ``Winner-Take-All" paradigm of network computation. The two concepts are classical but, to the best of our knowledge, they have always been studied separately. Rebound excitability is a {\it cellular} property that has long been identified as a core mechanism of rhythm generation \cite{brown_intrinsic_1911}. It has long been studied in neurophysiology and has inspired the design of artificial central pattern generators in rhythmic machines \cite{ijspeert_central_2008,ijspeert_simulation_2005,matsuoka_sustained_1985,matsuoka_analysis_2011}. Winner-Take-All computation is a {\it network} property at the core of decision-making mechanisms in computational biology and in computer science \cite{maass_computational_2000,usher_time_2001,lazzaro_winner-take-all_1988,neftci_synthesizing_2013, grossberg_adaptive_1976}. It has played a central role in the development of neuromorphic computation. A central message of this paper is to show that continuous rhythm generation and discrete decision making become one and same feature when combining the cellular property of rebound excitability and the network property of winner-take-all computation.

\begin{figure}[H]
    \centering
    \includegraphics[width=0.8\linewidth]{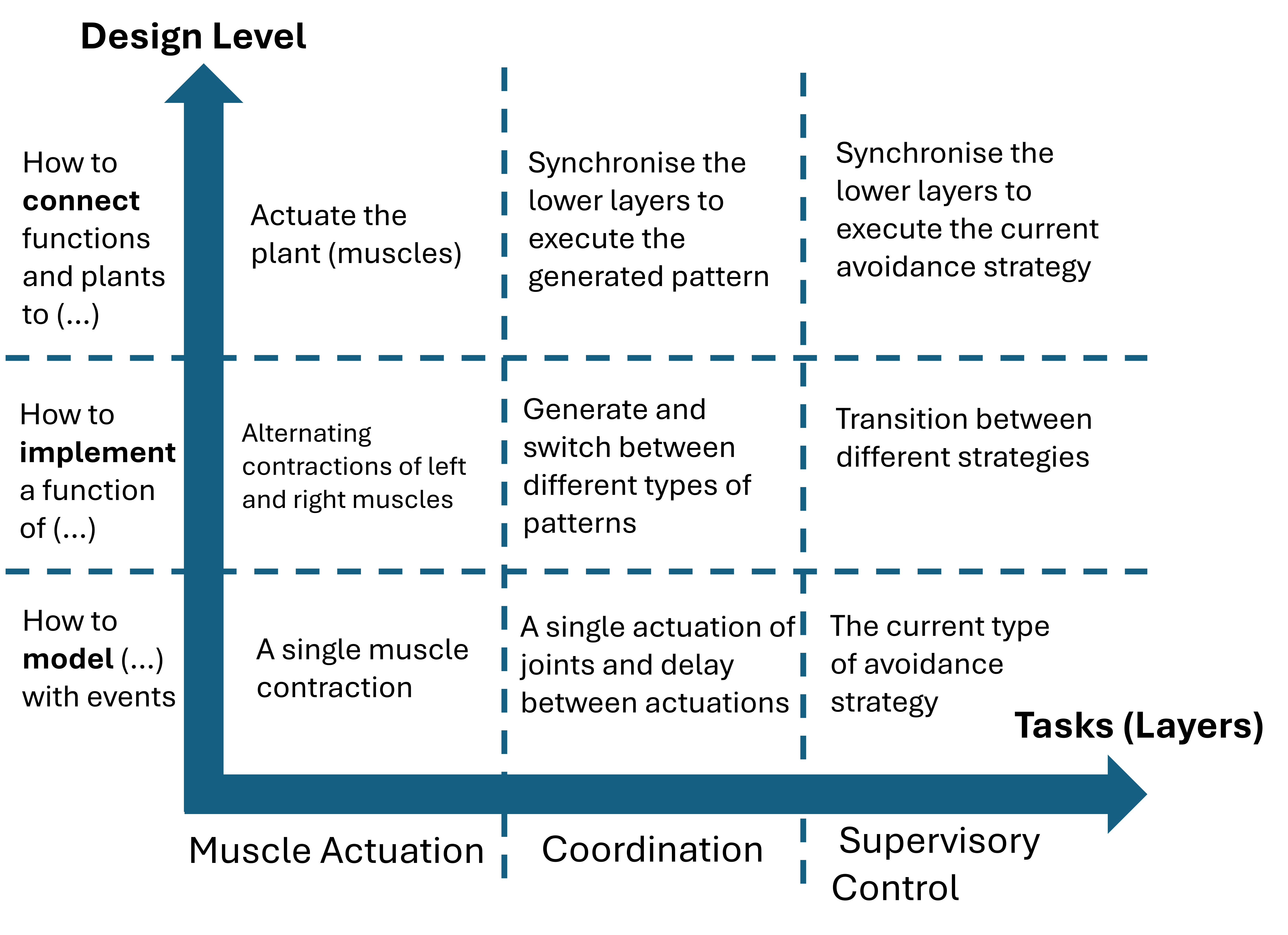}
    \caption{Design requirements for each design level in each task(layer).}
    \label{fig:intro}
\end{figure}

We illustrate this principle with the design of  a ``snake" nervous system via a multi-scale event-based neural network. Figure \ref{fig:intro} illustrates the conceptual hierarchical organization of the machine design. This organization is quite generic and the snake example is only chosen for the sake of a concrete illustration. Traditional control architectures would separate the use of physical models for the ``low-level" tasks (e.g. muscle actuation) from the use of computational models for the ``high-level" tasks (e.g. the supervisory control deciding to move straight or  turn left or right). Instead, our design uses the same neuronal architecture all the way from the single neuron to the entire snake controller. Through this unified architecture, we seek to mix ``regulation" and ``automation" at every stage of the design, so that the resulting event-based machine combines the best of the two worlds, namely the reliability of automation and the tunability of regulation. 

The resulting design is inherently modular, event-based, and multi-scale. The modularity stems from the bottom-up design of a hierarchical neural network made of simple functional motifs. The event-based nature of the network stems from the cellular rebound property of the elements. The multi-scale nature stems from the fact that the hierarchical organization of events: events at higher scales are made of ordered sequences of events at lower scales. Such a hierarchy necessarily results in a monotone spread of temporal and spatial scales from finer levels to coarser levels. Because the ``tuning" capabilities are always determined at the cellular level, the tuning capabilities are only limited by the finest resolution. Because the ``computing" capabilities stem from logical constraints imposed by the network interactions, the control architecture remains reliable at any scale. 

The limited scope of the present paper is to focus on the {\it controller} design only. We do not address the divide between the hardware and the software or between the digital and the analog. Those implementations choices can be made independently from the controller design. The controller is a neuronal circuit, but every part of this circuit can in principle be simulated in a digital computer or emulated in a physical device made of neuromorphic elements. 

The rest of the paper is organized as follows: Section~\ref{sec:event_level} reviews the cellular property of  rebound excitability in biophysical and neuromorphic circuits. Section~\ref{sec:decision_making} introduces the logic gates motifs and Winner-Take-All network. Section~\ref{sec:function-hco} introduces the rebound Winner-Take-All network and explains how different patterns can be generated with rebound Winner-Take-All network motifs. Section~\ref{sec:interaction_RWTA} describes how external events (pulses/biases) tune and interact with these motifs. Section \ref{sec:hierachy} extends the rebound Winner-Take-All network concept from network of cells to network of smaller networks. Section~\ref{sec:hierachical controller} composes the motifs into a five-link snake controller across muscle, pattern, and supervision layers. Discussions and Conclusions are in Sections~\ref{sec:discussion} and \ref{sec:conclusion}. 
All parameter details for all the plots are given in the appendix. Related code can be found in \url{https://github.com/Huoyongtony/A-Neuromorphic-Architecture-for-Scalable-Event-Based-Control}.

\medskip
\noindent\underline{Related work.}  The literature on rhythm generation and Winner-Take-All computation is vast. It is impossible to be exhaustive, but we survey a number of references that directly pertain to the work of the present paper. 

\subsubsection*{Rhythm generation}

Rhythm generation is a long standing question of  neurophysiology rooted in the pioneering concept of central pattern generator (CPG)
\cite{brown_intrinsic_1911,brown_nature_1914,wilson_central_1961}, whose  interaction with sensory feedback \cite{rossignol_dynamic_2006,
guertin_mammalian_2009} and neuromodulation
\cite{marder_central_2001,dirk_central_2015} generates robust rhythmic activity throughout nervous systems. Computational studies explain how reciprocal inhibition and post-inhibitory rebound (PIR) produce diverse oscillatory modes in the so-called Half-Center Oscillator (HCO)
\cite{skinner_mechanisms_1994,morozova_reciprocally_2022,matsuoka_analysis_2011}.
The biology of rhythms has inspired a lot of research in robotics and neuromorphic engineering, from the early adaptive CPG chip that powered an under-actuated leg \cite{lewis_toward_2000} to recent spiking implementations on a salamander robot that can swim and walk \cite{ijspeert_central_2008,ijspeert_simulation_2005,yu_survey_2014,
lozano_control_2016}, regulate a pendulum \cite{schmetterling_neuromorphic_2024}, ankle-rehabilitation
devices \cite{luo_preliminary_2024}, hexapods \cite{lopezosorio_neuromorphic_2024},
compliant manipulators \cite{abadia_neuromechanics_2025} and quadrupeds
with online CPG optimization \cite{zhang_online_2024}. Throughout this literature,  the design of the controller \emph{separates} system level  computation tasks  from CPG rhythm generation, connecting the two with ad-hoc glue logic
\cite{linares-barranco_towards_2022}.

\subsubsection*{Winner-Take-All computation and neuromorphic state machines}

Winner-Take-All (WTA) networks implement one of the core computational functions of analog computation: they perform analogue \emph{max} of \emph{softmax}  computation and serve as the “logic gates’’ of continuous-time computation \cite{majani_k-winners-take-all_1988,maass_computational_2000,
oster_computation_2009,grossberg_adaptive_1976}.

Early continuous-time Hopfield circuits demonstrated
$k$-WTA selection~\cite{majani_k-winners-take-all_1988}, and computational
complexity studies confirmed the expressive power of WTA
modules~\cite{maass_computational_2000}.
Indiveri and colleagues translated these ideas to analogue VLSI, proposing
compact soft-WTA arrays for low-power sensing and decision
\cite{indiveri_winner-take-all_1997,indiveri_artificial_2009,
indiveri_neuromorphic_2011}.
In spiking hardware, Oster \emph{et al.} showed rapid discrimination under
realistic input statistics~\cite{oster_computation_2009}, while
Linares-Barranco \emph{et al.} integrated a WTA core with CPGs on a mixed-signal
processor~\cite{linares-barranco_towards_2022}.
More recently, Cotteret \emph{et al.} embed finite-state machines in attractor-based
spiking networks using vector-symbolic arithmetic (VSA) to configure a
Hopfield weight matrix, with states encoded in \emph{rate} space
\cite{cotteret_vector_2024,cotteret_distributed_2025}.

Beyond hardware, a number of studies have enabled computational functions that are essential in human high-level cognitive systems to be implemented on spiking devices, such as finite state machines,
auto associative memories \cite{hopfield_neural_1982,abrahamsen_time_2004,indiveri_artificial_2009,cotteret_vector_2024,
cotteret_distributed_2025} and logic gates, to be directly implemented in recurrent spiking dynamics. 

\subsubsection*{Dynamical winner-take-all networks.} The concept of generating dynamical spatio-temporal patterns from winner-take-all principles also has a long history, best illustrated by the principle of  winnerless competition in neural dynamics \cite{rabinovich_dynamical_2001}. Winnerless competition circuits interconnect {\it rate} neuronal models  to build heteroclinic trajectories between saddle points, which raises challenges in ensuring both sensitivity to inputs and robustness against noise \cite{rabinovich_dynamical_2001,afraimovich_heteroclinic_2004}. A number of studies try to resolve this issue by determining the condition of robust heteroclinic \cite{rabinovich_discrete_2018,bick_dynamical_2009,seliger_dynamics-based_2003,ashwin_designing_2013,rabinovich_robust_2014,ashwin_criteria_2011} or adding external controllers \cite{herron_robust_2023}.  
Previous work has shown that winnerless competition controllers deliver tunable patterned motion, including snake locomotion \cite{horchler_designing_2015}.

The key difference between the mechanism of winnerless competition and the approach of the present paper is that we separate the {\it cellular} property of rebound excitability and the {\it network} property of Winner-Take-All computation. Rhythm generation in neuronal networks that have no excitability property at the cellular level requires an accurate tuning of the synaptic interconnections. In contrast, the restricted role of synaptic interactions in rebound-Winner-Take-All networks is to logically constrain the sequence of events. The decoupling between ``event generation" and ``event orchestration" is a key source of robustness and modularity in the proposed architecture.

\section{Cellular Level: Rebound Excitability, Neuronal Models, and Synapses. }\label{sec:event_level}

\subsection{Rebound Excitability}

The mechanism of rebound excitability has been studied in great detail since the early days of neurophysiology. We refer the reader to \cite{ribar_neuromorphic_2021} 
for a detailed survey of the relevant physiology as well as the neuromorphic question of designing artificial neurons with tunable rebound excitability, we refer the reader to \cite{ribar_neuromorphic_2021}. 

\begin{figure}[H]
    \centering
    \includegraphics[width=0.7\linewidth]{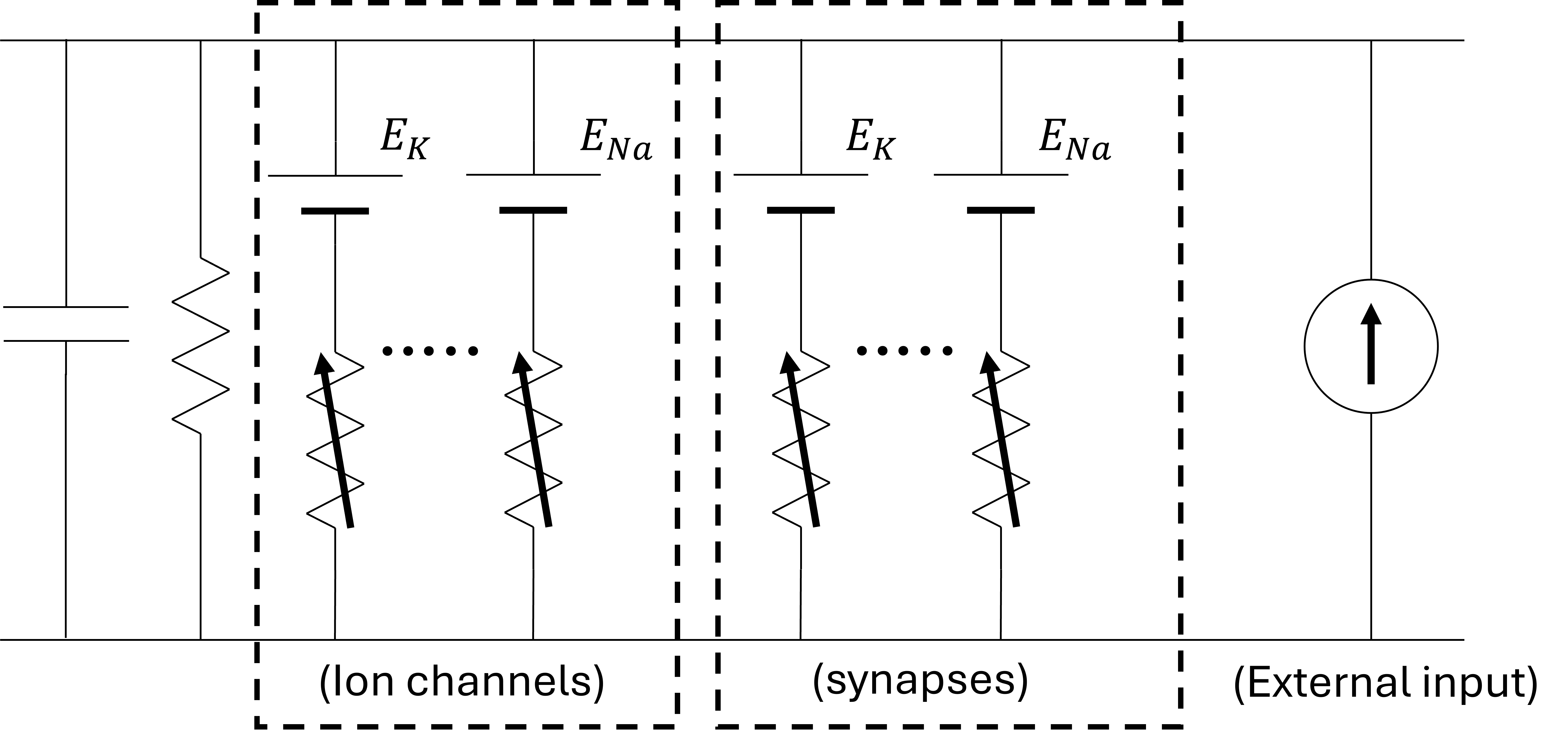}
    \caption{Circuit model of neuron}
    \label{fig:neuron circuit}
\end{figure}

In this paper, we rely on the classical circuit architecture laid out in the pioneering work of Hodgkin and Huxley \cite{hodgkin_quantitative_1952}.
A neuron is modeled as an electrical circuit that connects one leaky capacitor (the passive membrane) to a parallel bank of active filters defined by the series
combination of a battery and a memristive element, as shown in Figure \ref{fig:neuron circuit}. Each memristive element obeys the Ohmic law $i=gv$,  where the `conductance' $g$ depends nonlinearly and dynamically  on the membrane voltage. 

When the conductance depends solely on the internal capacitor voltage, $v$, the active filter represents an ion channel (e.g., sodium, potassium, calcium). This dependency defines an \emph{internal conductance}, regulating ion flow through the membrane  without external neuronal influence. If the conductance, however, depends on the membrane voltage of an external neuron, it is classified as an \emph{external conductance} and models a synaptic interconnection.
Whether internal (ion channel) or external (synapse), the precise model of a conductance $g$ varies across the literature. A conductance is often characterized by its activation and inactivation functions, each characterized by a specific voltage range and time-constant. 

\emph{Excitability} is the key property of a biophysical neuron \cite{sepulchre_excitable_2018}: the voltage response to a current stimulus is passive for small stimuli and {\it event-based} when the stimulus exceeds a threshold. The simplest event-based response of a neuron is a single spike.  An important property of biophysical neurons is that both depolarizing inputs (excitation - the input current flow increases the membrane voltage) and hyperpolarizing inputs (inhibition - the input current flow decreases the membrane voltage) can trigger an event. In the latter case, the event occurs at the end of the inhibition stimulus. Such an event,
illustrated in Figure \ref{fig:Simplified rebound spiking neuron}, is called a {\it rebound event}, and the corresponding excitability property is called {\it rebound excitability}. 

In conductance-based models, the simplest spike event is the result of the interaction of two currents: a fast-activation inward current (sodium channels, in the model of Hodgkin and Huxley) that creates positive feedback, and a slow-activation outward current (potassium channels, in the model of Hodgkin and Huxley) that repolarizes the neuron through negative feedback. 
Additional internal conductances can be used to control the excitability features of the neuron, including its sensitivity and robustness to dynamic uncertainties and perturbations. These conductances also enable the generation of specific event types. In this paper, we limit our discussion to spike and burst events at the cellular level. 

Rebound excitability is a crucial mechanism for sequentially propagating events through a chain of neurons connected by inhibitory synapses. An event in a presynaptic neuron creates transient inhibition in the postsynaptic neuron, which subsequently terminates in a rebound event. This basic mechanism enables the generation of reliable sequences of events. The duration and properties of each event are tuned by the internal conductances of the respective neuron; a cellular-level tuning mechanism. The overall sequence of these events is instead defined by the synaptic interconnection; a network-level tuning mechanism.

\subsection{Rebound Spiking and Bursting}
Throughout the paper, we focus on qualitative properties that do not depend on fine details of conductance models.
Therefore, for simplicity, we adopt the model proposed in \cite{ribar_neuromorphic_2021}. This model captures the essential properties of neurons, including \emph{rebound excitability}.

\begin{figure}[H]
    \centering

    \begin{subfigure}[t]{0.46\textwidth}
        \centering
        \includegraphics[width=\textwidth]{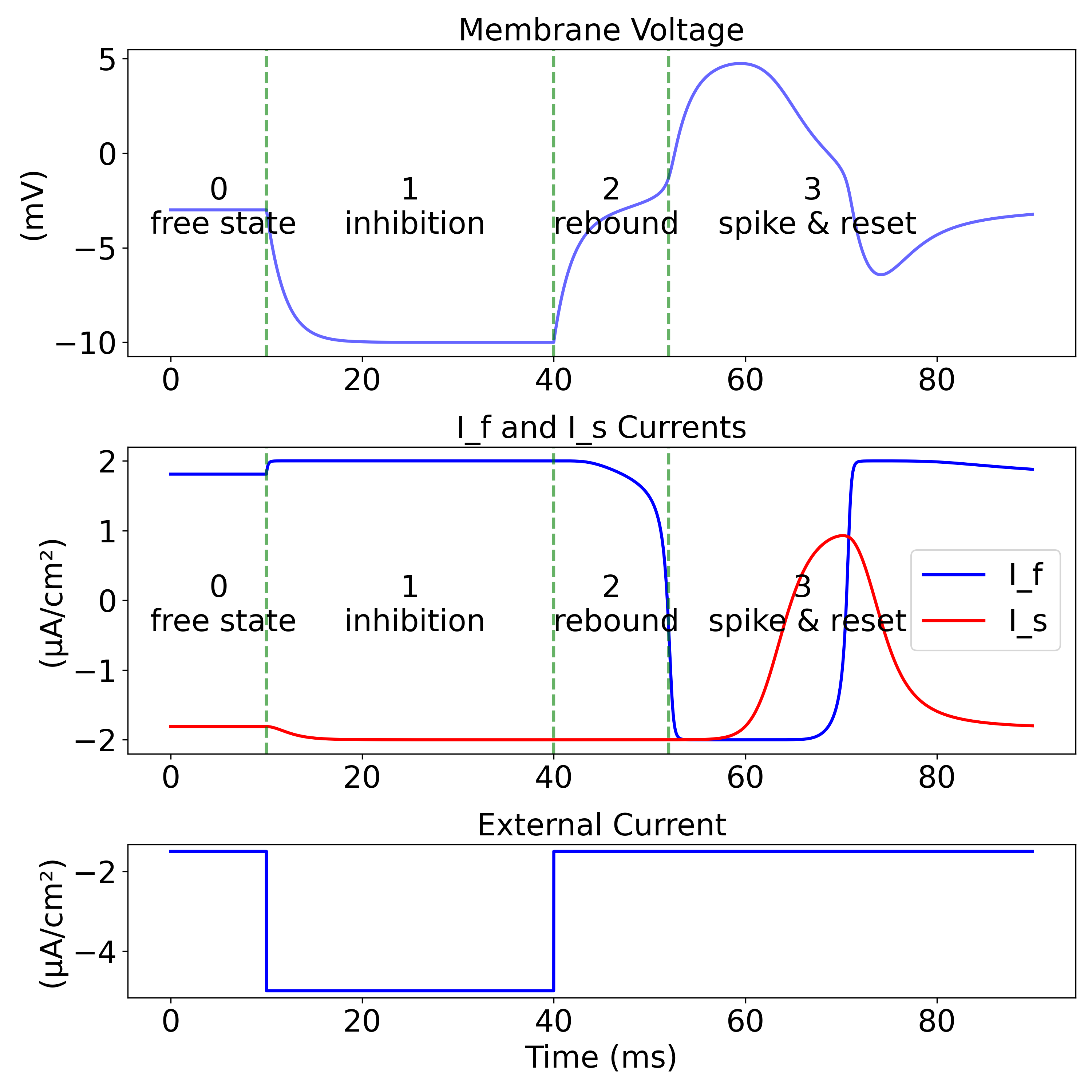}
        \caption{Rebound spiking neuron.}
        \label{fig:Simplified rebound spiking neuron}
    \end{subfigure}
    \hfill
    \begin{subfigure}[t]{0.46\textwidth}
        \centering
        \includegraphics[width=\textwidth]{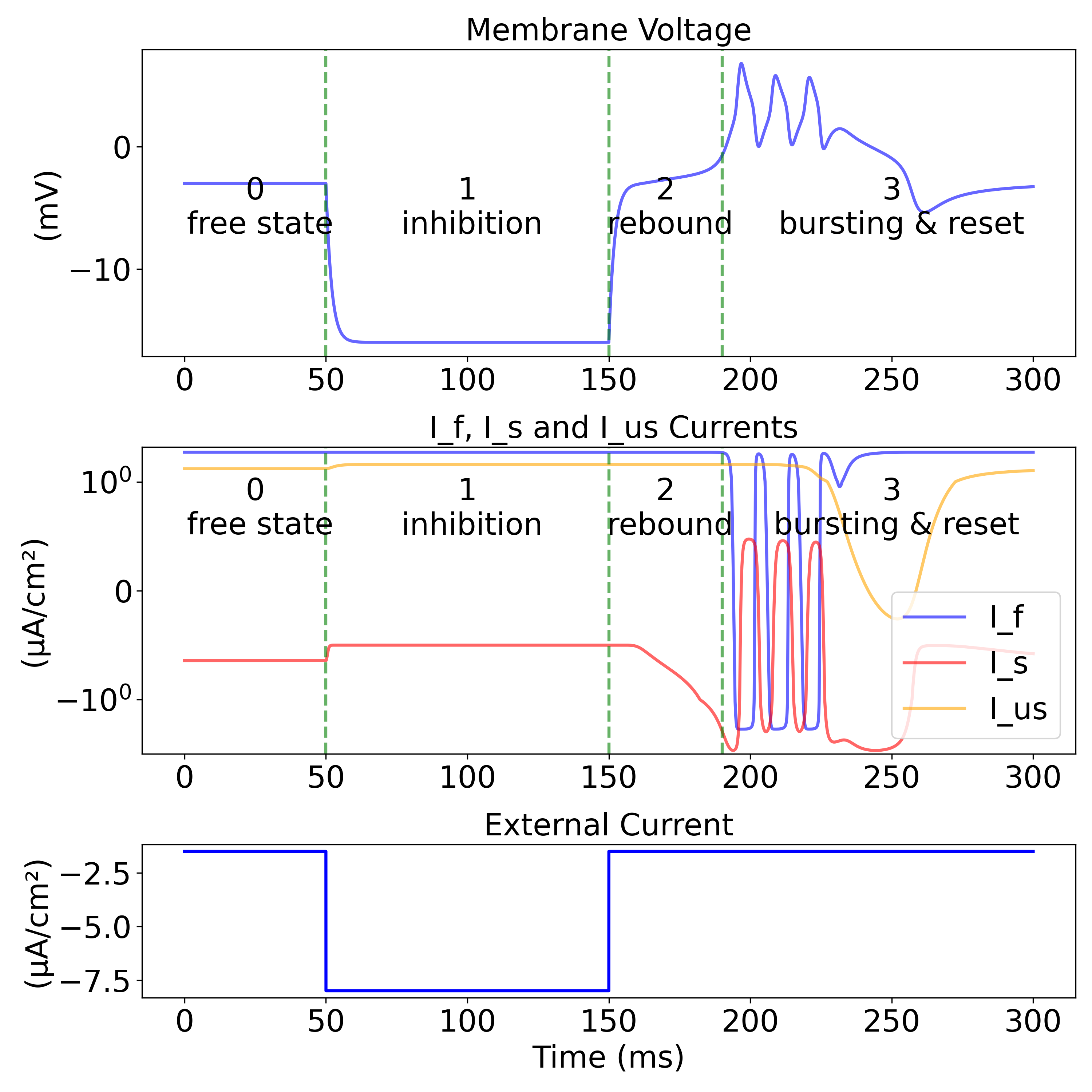}
        \caption{Rebound bursting neuron.}
        \label{fig:Simplified rebound bursting neuron}
    \end{subfigure}
    \caption{Rebound neurons.}
    \label{fig:main}
\end{figure}

As shown in \eqref{eq:rsn}, the  membrane voltage (RC circuit) of a \emph{rebound spiking neuron} is driven by the external current $I_{\mathrm{ext}}$ and by two internal
currents: a fast depolarizing current $I_f$ (static nonlinearity) and a slow repolarizing current $I_s$ (cascade of low-pass filter and static nonlinearity).
The parameters  $\alpha$s regulate the strength of the currents.
$\bar V$ models the presence of batteries.  The rebound spiking excitability of the neuron is illustrated in Figure  \ref{fig:Simplified rebound spiking neuron}. 
\begin{equation}
\label{eq:rsn}
\begin{aligned}
    C\dot{V}&= -\frac{1}{R}V - I_{f} - I_{s} + I_{\rm ext} \\[2mm]  
    I_{f}&= -\alpha_f^{-}\tanh(V-\bar V) \\
    I_{s}&= \alpha_s^{+}\tanh(V_s-{\bar V}) \\
    \tau_s \dot{V}_s & = -V_s+V  
\end{aligned}
\end{equation}


In a similar way, the mathematical model of a \emph{rebound bursting neuron} is given \eqref{eq:rbn}, characterized by the introduction of a new pair of currents, at a slower time scale (an additional term within the
definition of $I_s$ and the ultra-slow current $I_{us}$). This is the time-scale of
the burst, a cluster of of spikes separated by periods of quiescence (no firing). The role of these additional terms is to produce a slower overshoot, similar to a long smoothed spike, which then turns into a burst. The rebound bursting excitability of the neuron is illustrated in Figure \ref{fig:Simplified rebound bursting neuron}, which shows the membrane voltages and the ion currents at 4 stages of rebound excitation. 

\begin{equation}
\label{eq:rbn}
\begin{aligned}
    C\dot{V}&= -\frac{1}{R}V - I_{f} - I_{s} - I_{us}+ I_{\rm ext} \\[2mm]  
    I_{f}&= -\alpha_f^{-}\tanh(V-{\bar V}_f) \\
    I_{s}&= \alpha_{s}^{+}\tanh(V_s-{\bar V}_f) - \alpha_{s}^{-}\tanh(V_s+{\bar V}_s) \\
    I_{us}&= \alpha_{us}^{+}\tanh(V_{us}-{\bar V}_s) \\
    \tau_s \dot{V}_s & = -V_s+V  \\
    \tau_{us} \dot{V}_{us} & = -V_{us}+V  
\end{aligned}
\end{equation}

\begin{figure}[H]  
    \centering
    \includegraphics[width=0.46\textwidth]{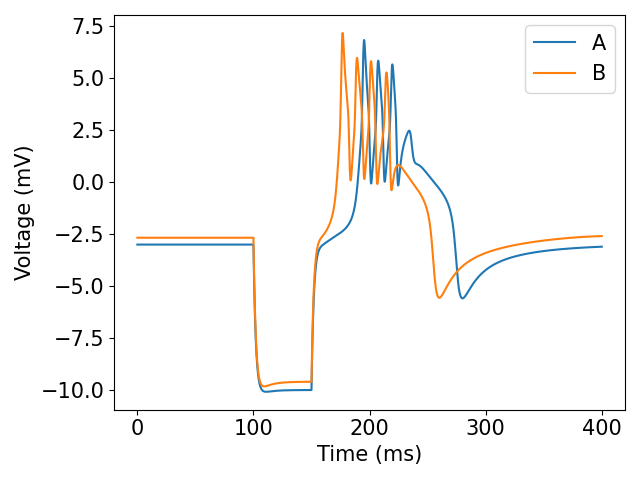}
    \caption{Two voltage trajectories, A and B, of the same neuron for different values of $\alpha_{us}^{+}$. The input current is an inhibitory step from -1.5 to -5 with duration 50 ms. Trajectory A (B) corresponds to a smaller (larger) $\alpha_{us}^{+}$.}
    \label{fig:bursting_neuron_2}
\end{figure}

As before, the parameters  $\alpha$s regulate the strength of the currents and ${\bar V}_f$ and ${\bar V}_s$ model the effect of batteries. As shown in \cite{ribar_neuromorphic_2021}, a burst rebound event is similar to a spike rebound event but it has a
well-defined and tunable duration. The burst duration is regulated by the ultra-slow parameter $\alpha^{+}_{us}$, as shown in Figure \ref{fig:bursting_neuron_2}. Hence we can regard a spiking event as an event of \emph{fixed} duration and a bursting event as an event of \emph{variable} duration.
%
%
%

\subsection{Inhibitory and Excitatory Synapses} 

Synaptic currents model directional interconnections 
between neurons. A synaptic current is (dynamically) determined by the presynaptic neuron membrane voltage and affects the postsynaptic neuron membrane voltage.
In this paper we adopt the simplified synaptic model  of \cite{ribar_neuromorphic_2021}. A synaptic current $I_{ij}$ of type $k$ from the presynaptic neuron $j$ (voltage $V_j$) to the postsynaptic neuron $i$ has the expression 
\begin{equation}
\label{eq:syn}
\begin{aligned}
    I_{ij} &= G_{ij}\sigma(V_{ij}-\bar{V}_{ij})^k \\ 
    \sigma(V_{ij}-\bar{V}_{ij})&= \frac{1}{1 + e^{-\alpha_{ij} (V_{ij}-\bar{V}_{ij})}}\\
    \tau_{ij} \dot{V}_{ij} &= -V_{ij} + V_j,
\end{aligned}
\end{equation}
where $G_{ij}$ is the synaptic weight,
$\bar{V}_{ij}$ represents the contribution of the battery, and
$\alpha_{ij}$ regulates the slope of the (sigmoidal) function $\sigma(\cdot - \bar{V}_{ij})$ in the neighborhood of
$\bar{V}_{ij}$. As for the case of internal currents, synaptic currents are modeled as the cascade of a low-pass filter with time constant $\tau_{ij}$ and a biased sigmoidal nonlinearity. 
The synaptic weight  $G_{ij}$ is \emph{positive} for \emph{excitatory} synapses and \emph{negative} for \emph{inhibitory} synapses. 

The synaptic model of \eqref{eq:syn} simplifies the classical Ohmic model of a physiological synapse. The latter has an additional multiplicative factor that determines the sign of 
the current. In contrast, in \eqref{eq:syn}, 
the excitatory or inhibitory nature of a synapse is 
fully determined by the sign of $G_{ij}$; an easier characterization.
The synapse can be thought of as a nonlinear filter on the presynaptic voltage. The quality of this filtering action provides different `readouts' of presynaptic events. For example, a fast synapse with a high $\bar{V}_{ij}$ driven by a bursting neuron will produce a cluster of current pulses (several events in current space). In contrast, a slow synapse with a low $\bar{V}_{ij}$ will low-pass filter the cluster of spikes into a slowly rising and decaying current bump (a single event in current space).

In this paper, all synapses are confined to three time-scales:
fast ($\tau \simeq 0.5$), slow ($\tau \simeq 50$), and ultra-slow ($\tau \simeq 100$). In all figures of this paper, fast synapses are represented with blue, slow synapses are represented with red and ultra-slow synapses are represented with black. We use an \emph{arrow} for \emph{excitatory} synapses and a \emph{circle} for \emph{inhibitory}  synapses.

\section{Functional Level: Decision Making}\label{sec:decision_making}
\begin{figure}[H]
    \centering
    \includegraphics[width=0.9\linewidth]{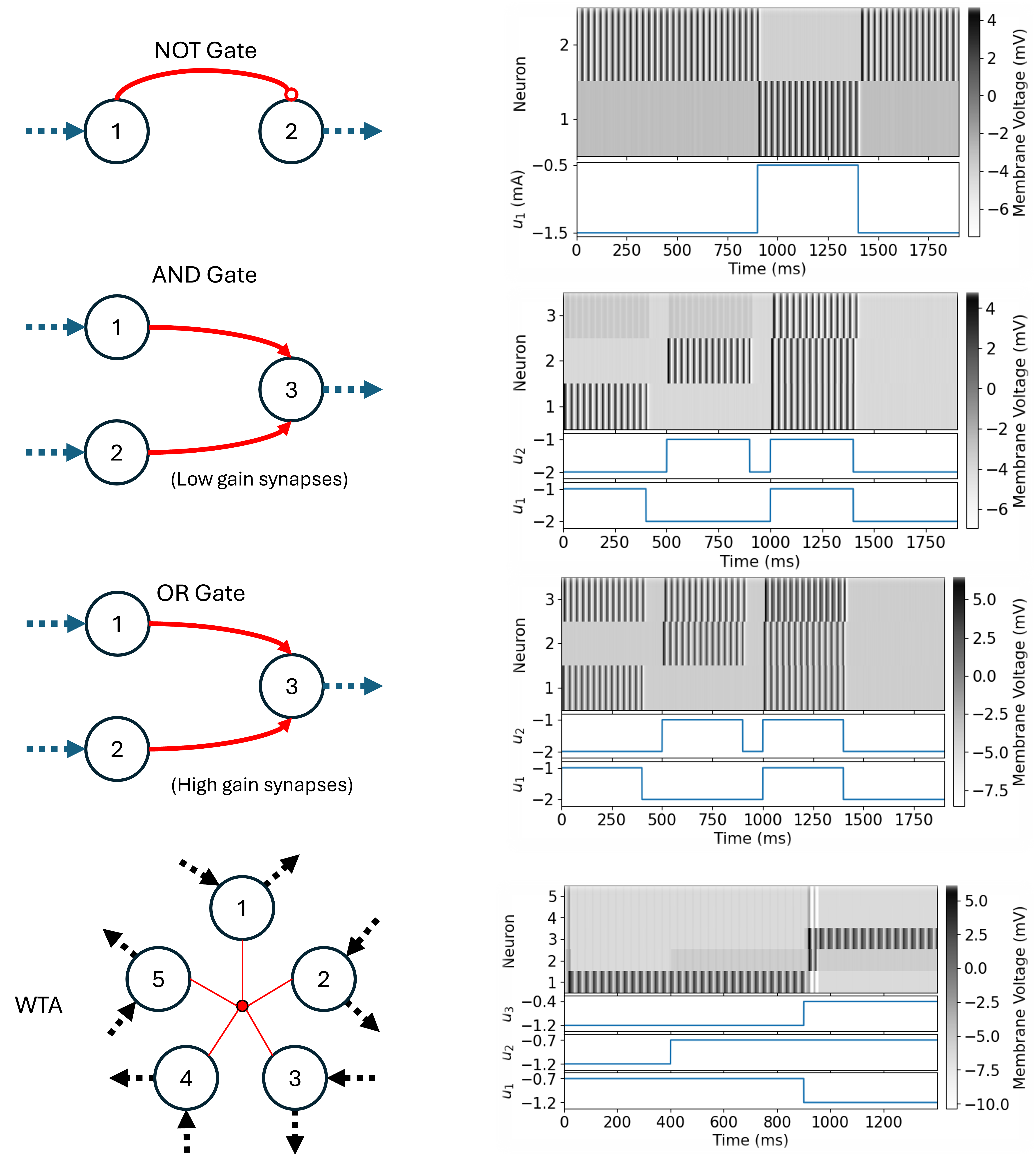}
    \caption{Decision Making Motifs. Fast synapses ($\tau=0.5$) are represented with blue, slow synapses ($\tau=50$) are represented with red. Arrow for excitatory connection and circle for inhibitory connections. Dashed arrows denote input and output connections.}
    \label{fig:decision_making}
\end{figure}

Logical operations such as NOT, AND and OR are the fundamental elements of decision making. In this section, we provide a neuromorphic realization of these classical logical elements. We then move to a more complex network, 
illustrating a neuromorphic realization of the classical Winner-Take-All mechanisms (WTA) \cite{grossberg_adaptive_1976}. Logic gates and winner-take-all mechanism realizations are illustrated in 
Figure \ref{fig:decision_making}.
\begin{description}
\item[\textbf{NOT gate:}]
A single inhibitory synapse implements a NOT gate. 
The NOT function means that
an event in the presynaptic neuron forbids a simultaneous event in the postsynaptic neuron. 
\item[\textbf{AND gate:}] Two rebound neurons (input nodes) connect with excitatory synapses to a third rebound neuron (output node). The maximum gain of synapses can be chosen such that events in the two presynaptic neurons are necessary to trigger an event in the postsynaptic neuron. 

\item[\textbf{OR gate:}] The architecture is similar to the AND gate. However, the maximum gain of the synapses is increased to guarantee activation of the postsynaptic neuron whenever either one of the two presynaptic neurons spikes. 
\end{description}
In digital electronics, OR (AND) gates can also be realized using NOT and AND (OR) gates. Similar constructs can be proposed in our neuromorphic setting. These derived neuromorphic circuits would exhibit longer propagation delays but offer interesting features in terms of robustness and tunability of their response to a stimulus. For instance, the OR gate proposed in Figure \ref{fig:decision_making} shows sensitivity to the selection of synapses weights. In contrast, though consuming more energy and introducing greater delays, its realization through NOT and AND neuromorphic circuits offers reduced sensitivity to the specific choice of synaptic gains in the network. \vspace{2mm}

\begin{figure}[H]
    \centering
    \includegraphics[width=0.5\columnwidth]{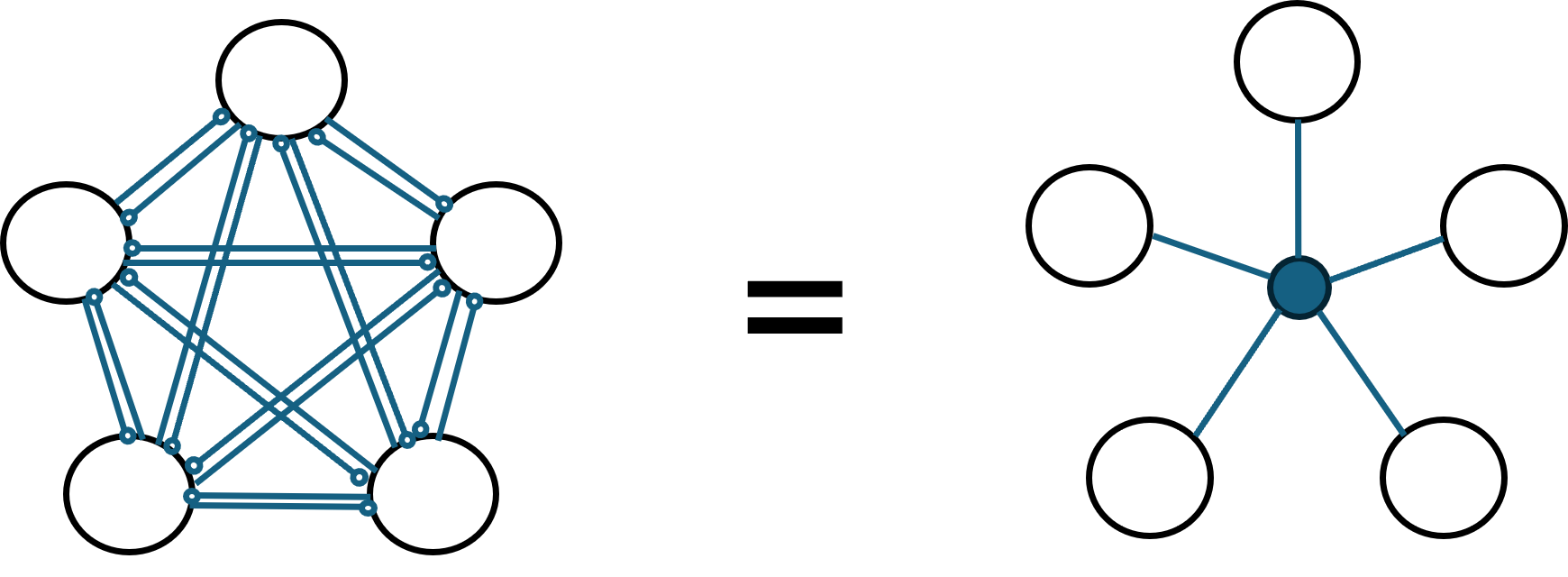}
    \caption{Equivalent representations of an all-to-all inhibitory network.}
    \label{fig:inhibition}
\end{figure}

\begin{description}
\item[\textbf{Winner-Take-All (WTA) network:}]
 This is a competitive neural circuit that compares multiple inputs and amplifies the strongest one while suppressing the rest, typically through lateral inhibition. Functionally, it converts small differences in input strength into a clear selection. 
Winner-Take-All network is typically an all-to-all inhibitory network,
as shown at the bottom of Figure \ref{fig:decision_making}. To avoid clutter, the all-to-all connectivity is represented by a central node (see also Figure \ref{fig:inhibition}).
\end{description}
The raster plot on the right of the WTA network in Figure \ref{fig:decision_making} illustrates its input-output behavior. The
inputs $u_1$-$u_3$ are the currents injected into neurons below 1-3.Neuron 1 receives the highest input at the start while all other neurons are simply affected by noise. Hence, neuron 1 becomes the winner. Even though the input of neuron 2 is later increased, the selected `winner' neuron does not change. This demonstrates the hysteresis behavior at the core of the Winner-Take-All network. The network becomes again sensitive to external inputs only when the input to neuron 1 is reduced. As shown in Figure \ref{fig:decision_making}, in such a case a slightly higher input to neuron 3 than to neuron 2 sets the new winner (neuron 3).  

\section{Functional Level: Pattern Generation}\label{sec:function-hco}
Unlike pure logic operations, events in the real world unfold in time, and neural circuits respond to them with specific time dynamics. This dynamic time response is crucial to shape how the circuit interacts with its external environment. In this section, we explore this temporal dimension
by introducing rebound neural networks for pattern generation. 
We proceed from the simplest case, the Half–Center Oscillator, to the more general, the Rebound Winner–Take–All network, showing how rebound at the cellular level and inhibitory competition at the network level combine to yield robust spatio–temporal patterns.

\begin{figure}[H]
    \centering
    \includegraphics[width=0.9\textwidth]{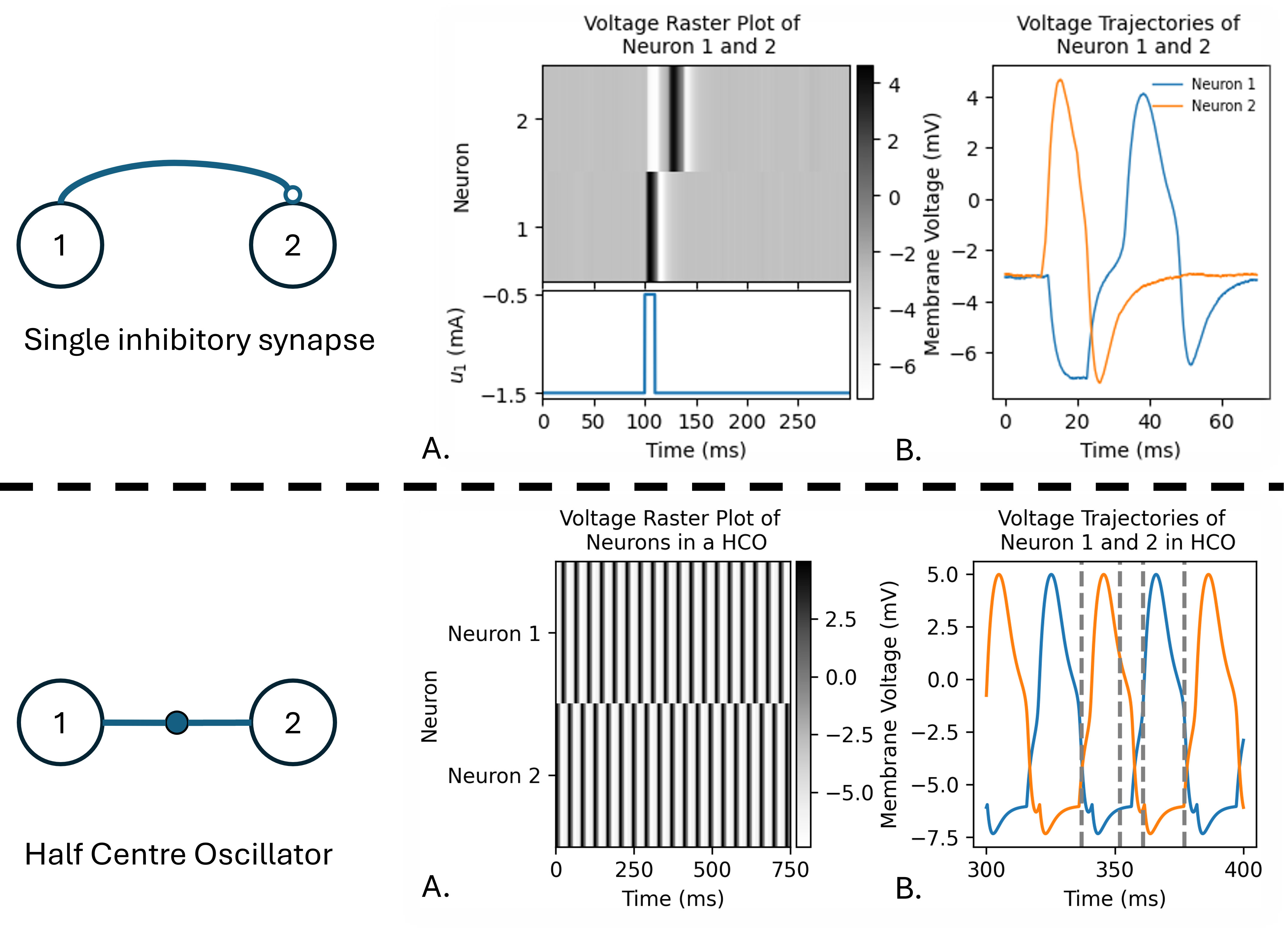}
    \caption{The half-center oscillator is the simplest example of a RWTA network. The two neurons of the network rebound in alternance.}
    \label{fig:HCO_pattern}
\end{figure}

\begin{description}
\item[\textbf{Half–Center Oscillator (HCO):}]
It consists of two rebound neurons reciprocally connected by inhibitory synapses, as shown at the bottom row of Figure\,\ref{fig:HCO_pattern}.
It is the simplest pattern generator built from 
rebound neurons \cite{brown_intrinsic_1911}.
\end{description}

A useful way to understand the network is to start with a \emph{single} inhibitory synapse from neuron~1 to neuron~2 (Figure\,\ref{fig:HCO_pattern}, top row). When neuron~1 spikes, it inhibits neuron~2; upon release of inhibition, neuron~2 produces a rebound spike. With \emph{reciprocal} inhibition, this rebound triggers the reverse action, yielding an alternating left–right pattern. This rebound–based mechanism is classical and extensively documented \cite{marder_central_2001,dirk_central_2015}.

In our implementation, the synapse dynamics for the HCO use \(\tau = 0.5\), \({\bar V}=-1\), \(\alpha=50\), and weight \(-2\). The threshold is chosen near or below the neuron's spiking threshold so that the postsynaptic neuron activates only when the presynaptic neuron has returned to rest; \(\tau\) is fast enough to preserve the spike after filtering; \(\alpha\) is large enough to saturate during rebound. The resulting behavior is shown in Figure\,\ref{fig:HCO_pattern}.

\begin{description}
\item[\textbf{Rebound Winner-Take-All (RWTA):}]
It consists of 
network of synaptically interconnected rebound neurons characterized by  the presence of fast all–to–all inhibitory connections to enforce that at most one neuron can spike at any time (the \emph{single–winner} constraint). The Rebound Winner-Take-All network generalizes the functionality of the HCO to networks of rebound neurons of arbitrary size. Figure \ref{fig:pattern_generation} illustrates 3 examples of RWTA motifs.
 They are: a ring oscillator of spiking neurons, a ring oscillator with spiking and bursting neurons, and a dual ring oscillator.
\end{description}

 The topology of these networks is based on two complementary principles:
 (i) \emph{Fast inhibition} for `state restriction'. The all–to–all inhibition limits the event state space to \(n\) states for \(n\) neurons. At any time, exactly one neuron can be `on' while the rest must be `off'. (ii) \emph{Slow excitation} to `enable specific transitions'. Directed slow excitatory couplings bias the competition so that, at the end of an event, a designated postsynaptic neuron is favored as the next winner. The time-scale separation between inhibition and excitation is a key feature. Excitatory synapses are \emph{slow} relative to inhibitory synapses so that the bias persists through the event and into the rebound stage of the next neuron. 
The following networks show how to construct specific temporal batterns and also illustrate the flexibility of the Rebound Winner-Take-All network.

\underline{Spiking Ring Oscillator:}
A ring of \(N\) spiking neurons with fast all–to–all inhibition and slow excitation connections forming a ring produces an autonomous traveling wave \(1\!\to\!2\!\to\!\cdots\!\to\!N\), in the absence of external input. The length of the inter spike interval is given by the combined contribution of spike interval and rebound interval (see Figure \ref{fig:Simplified rebound spiking neuron}). The former is mostly
affected by the time constant of the fast inhibitory synapses. The 
latter is sensitive to the time constant and weight of the slow excitatory synapses and it also increases monotonically with the input magnitude. More details can be found in \cite{huo_winner-takes-all_2025}.

Notably, for the case of two neurons $N=2$, the RWTA network corresponds to the Half-Center oscillator.
This also shows that the addition of excitatory connection on the HCO network does not violate its Winner-Take-All characteristics.

\underline{Spiking and Bursting Ring Oscillator:}
The features of the RWTA network are robust to heterogeneity. Replacing a spiking neuron with a bursting neuron  does not affect the pattern generation abilities of the network. The effect of the bursting neuron is to stretch inter–wave waiting interval while preserving the same sequence of events. Only the differences in intrinsic time-scales and thresholds might need to be compensated by neuron-specific bias currents and appropriately matched synaptic time
constants. In our simulation, we do not introduce
any compensation, that is, the interconnection parameters are kept constant between the spiking and the bursting cases. This shows one of the advantages of decoupling the network design from the cellular design. The topology of the network only controls the sequencing of events, but the events themselves are neuronal properties. This decoupled role of the network topology and the neuronal parameters is what makes the network topology robust to cellular heterogeneity. 

\underline{Multiple Ring Oscillators as Pattern Selector:}
This example demonstrates how the synapses determine the rhythmic properties of the network. Specifically, connecting two rings that \emph{share a common inhibitory hub} turns the network into a state machine. This architecture enables the network to select between two different patterns based on the external input. Because the hub enforces single–winner behavior across rings, a brief external pulse to any neuron can switch the active pattern. Figures~\ref{fig:pattern_generation} illustrates toggling between two rhythmic patterns, demonstrating discrete state transitions on top of rhythmic generation.

Across all these examples, we can identify
the following general properties of RWTA networks:
\begin{enumerate}[label=\textbf{P\arabic*}, leftmargin=2em]
\item \textbf{Single–winner:} at any time, at most one neuron is in an event state (the winner).
\item \textbf{Refractoriness:} events end with a refractory period, preventing a neuron from winning twice in rapid succession.
\item \textbf{Competitive selection:} during rebound, all neurons compete; slow excitation and/or depolarizing inputs designate the next winner.
\item \textbf{Event–driven sequencing:} cycles of slow excitation yield endogenous cyclic event sequences (e.g., ring waves).
\end{enumerate}
The specific separation of inhibition and excitation is at the core of these properties.
In the RWTA architecture, the design of the states (governed by fast inhibition) is effectively decoupled from the design of the transitions (governed by slow excitation). This qualitative interconnection design, which relies on topology and time-scales rather than on precise values, ensures robustness against parameter variability and neuron mismatch.

We complete this section by emphasizing that the computational principles of our neuromorphic architecture are consistent with the biophysical architecture of EI networks studied in neuroscience \cite{wilson_excitatory_1972, wilson_spikes_1999}. An EI network of rebound neurons was previously studied in  \cite{drion_switchable_2018} to show how cellular neuromodulation can easily reconfigure the network {\it states}. A neuromorphic neuron in our architecture corresponds to a pair of excitatory and inhibitory neurons in EI networks.  The inhibitory layer must be densely connected to approximate the all-to-all topology of a rebound network, and the inhibitory synapses are typically of the GABA type. In contrast, the excitatory layer is sparsely connected by AMPA synapses. While AMPA synapses are normally faster than GABA synapses, our  {\it slow} excitatory synapse can be regarded as a model of the synaptic delay resulting from event propagation through several nodes in the excitatory layer.

\begin{figure}[H]
    \centering
    \includegraphics[width=0.9\textwidth]{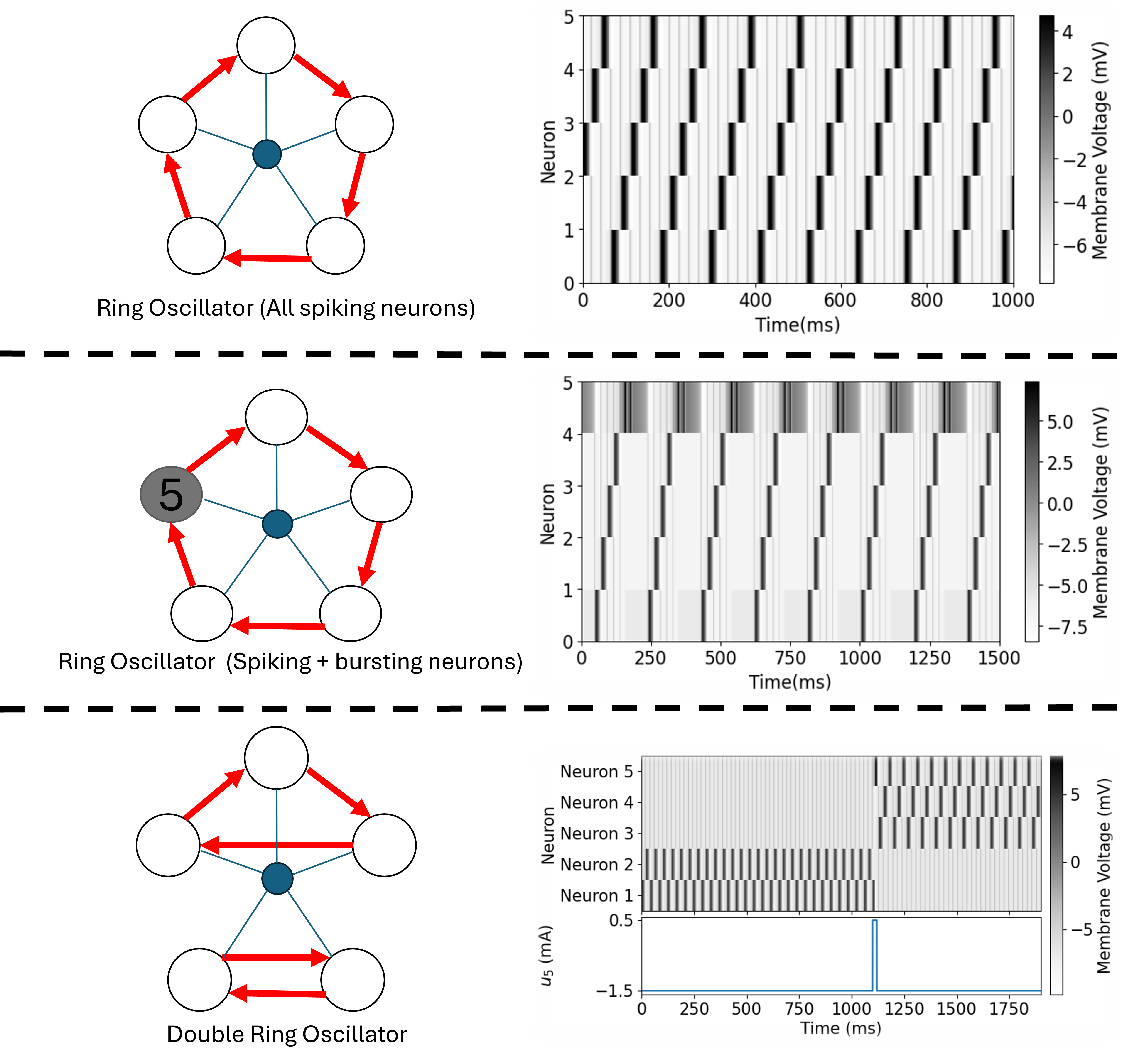}
    \caption{Rhythmic states generated by slow cyclic excitatory connections in a  RWTA network.}
    \label{fig:pattern_generation}
\end{figure}

\section{Functional Level: Interaction of the RWTA Network} \label{sec:interaction_RWTA}
\begin{figure}[H]
    \centering
    \includegraphics[width=0.9\linewidth]{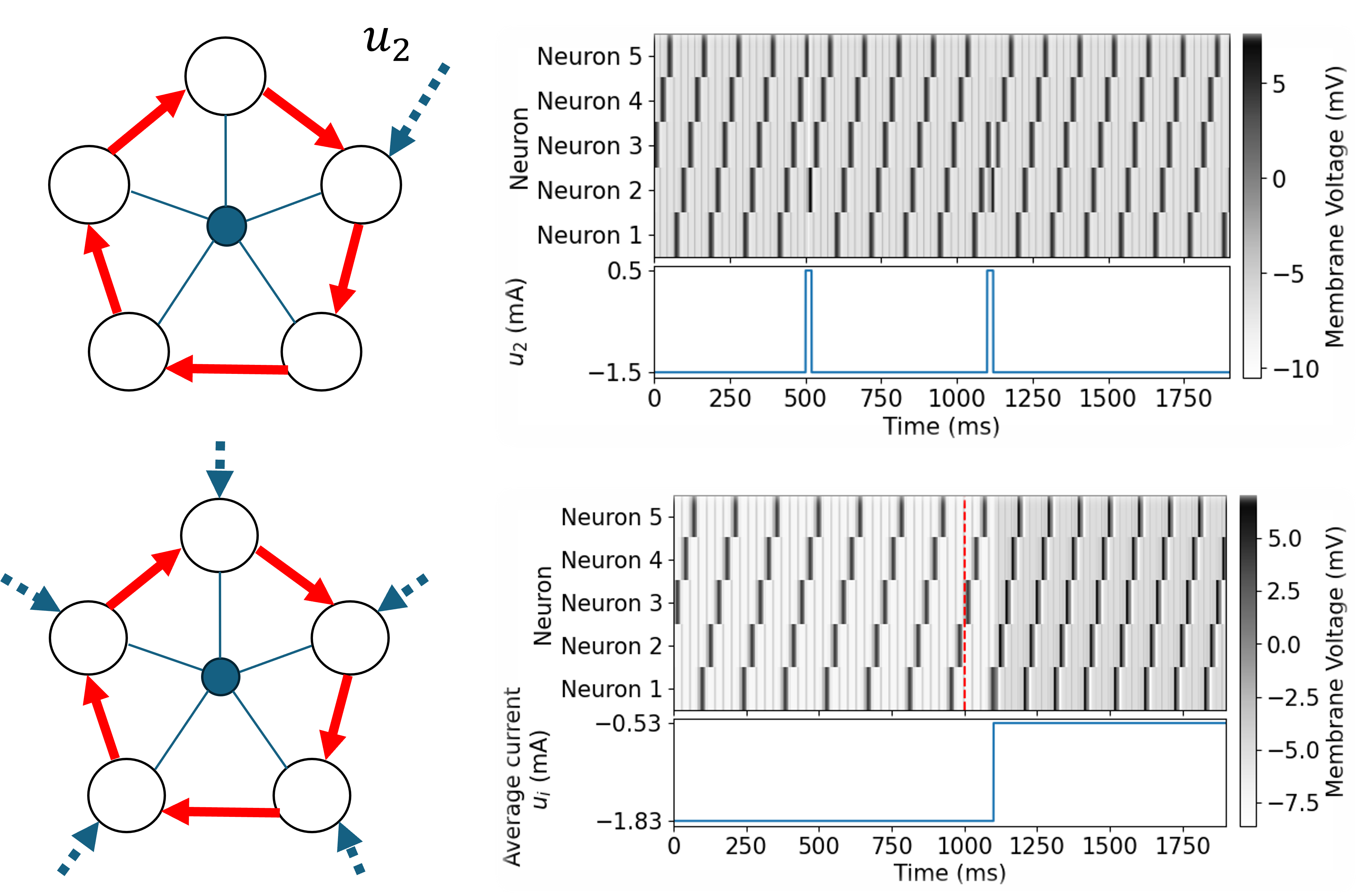}
    \caption{Two distinct mechanisms of external control: fast and localized external events can phase advance or reset the endogenous pattern; slow and global external inputs modulate global properties of the rhyhtm, e.g. its frequency.}
    \label{fig:Tuning_and_interaction}
\end{figure}

The rebound-Winner-Take-All architecture encodes reliable endogenous {\it states} (i.e. event sequences that do not require external input), but its {\it rebound} nature makes it also eminently interactive with the external environment. In the absence of external inputs, the rhythm is endogenous, that is, only determined by the internal network topology. But external inputs can either re-orchestrate or tune the rhythm as follows:

\underline{Phase switching ({\it local external interaction}):}
A \emph{brief pulse} $u_i(t)$ applied to neuron~$i$ during its rebound stage
raises its membrane voltage $V_i$, relative to its competitors.  If the pulse arrives before any
neuron has crossed threshold, $i$ wins the current competition and the network
\emph{jumps} to phase~$i$. If the pulse arrives during inhibition, the neuron will have a higher initial voltage at the start of next rebound and hence will increase the probability to win the next round. Thus, the RWTA network functions as an event-driven
Winner-Take-All selector that can be either hard-reset or synchronized to external events. 
This is illustrated in  Figure \ref{fig:Tuning_and_interaction}. Neuron 2 receives pulses at two different times and each time the network pattern is reset to Neuron 2.

\underline{Frequency modulation ({\it global external control}):}
A \emph{slow modulation of the common bias}
represented by $u_i\mapsto u_i+\Delta u$ for all neurons $i$, increases the rebound speed of
\emph{every} neuron. This stretches or compress the rebound interval of each neuron while leaving the firing order unaffected. 
As a result, the RWTA network
behaves as a current-controlled rhythmic pattern generator whose frequency
can be tuned continuously either in open loop (constant $\Delta u$) or in closed loop ($\Delta u$ is continuously regulated by a feedback loop).
Figure \ref{fig:Tuning_and_interaction} shows how a hyperpolarization step change in the the average current input decreases the frequency of the generated rhythm.

These two cases show how the
pattern generation capabilities of the RWTA network exhibit features from both discrete localized event-based interaction and continuous global neuromodulation behaviors, resulting in a versatile yet robust pattern generator.

\section{Multiscale States and Network Hierarchy}\label{sec:hierachy}
In this section we show that the rebound Winner-take-all network is hierarchical and multi-scale: neurons generate events, networks generate patterns of events that can themselves be regarded as new events at a broader scale, and so on. We limit our discussion to a few specific but conceptual examples.

\begin{figure}[H]
    \centering
    \includegraphics[width=1\linewidth]{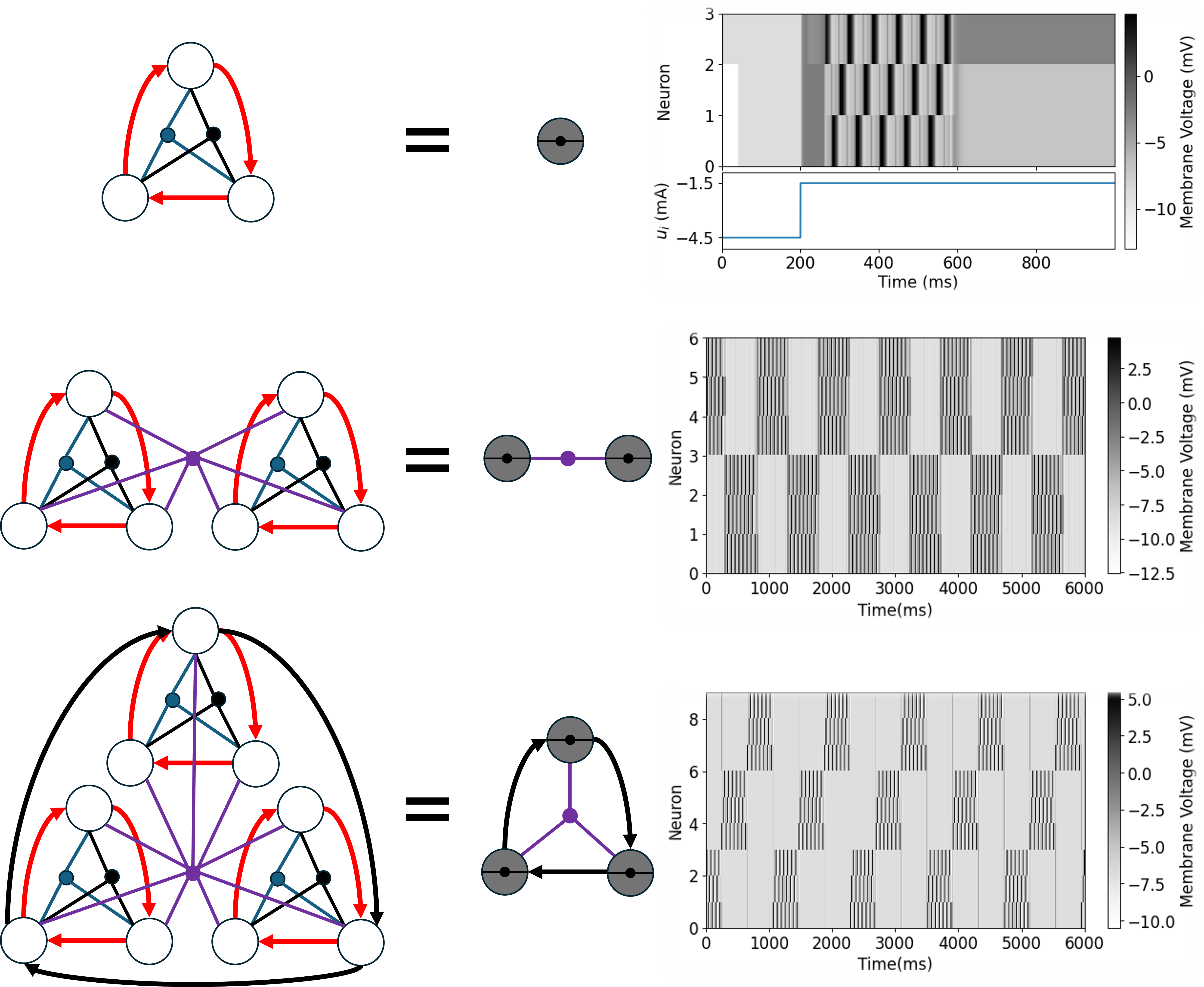}
    \caption{The RWTA architecture is inherently hierarchical, owing to the interlacing of E and I connections and the monotone ordering of time-scales.}
    \label{fig:hierarchy}
\end{figure}

Figure \ref{fig:hierarchy} shows the architecture of a  RWTA of RWTA networks. The top row represents the structure and simulation of a three nodes RWTA network. Rhythmic sequences are turned into events (of events) by adding ultra-slow all-to-all inhibitory synapses that control that terminate the cyclic pattern after a fixed duration. The effect of the ultra-slow inhibitory connections is illustrated in the top row simulations of Figure \ref{fig:hierarchy}.
All the neurons are initially driven by a strong external inhibition, released at, which is released at $200$ms. 
Right after the release, the network rebounds, generating a rhythmic pattern.
In contrast to the previous section, the pattern terminates after a few iterations. In other words, the pattern itself becomes a new event made of a sequence of events at finer scale.

In the second row of Figure \ref{fig:hierarchy} 
we have constructed a HCO of rebound networks. 
The two rebound networks are connected by slow all-to-all synapses, represented by a purple node. Their time scale is faster than the slow excitatory synapses (red) but can be slower or equal to the time scale of the fast inhibitory synapses internal to each network (blue). The simulation illustrates a clear rhythmic alternating pattern between the patterns of each rebound networks. 

As last example, we create a three-node RWTA 
network of RWTA networks, as shown in the bottom row of Figure \ref{fig:hierarchy}. The outer layer of excitatory connections between each node (black) are ultra-slow synapses with the same time constant as the all-to-all ultra-slow inhibitory synapses internal to each node (also in black). The simulation illustrate the pattern of a 3-ring oscillator of 3-ring oscillators.

These examples suggest that the design of a RWTA network is inherently hierarchical. Specific sequences of events form new events. With the same interconnection rules, those new events can be sequenced to create new patterns, etc.. The hierarchy of time-scales and the complementary role of excitation and inhibition are key to enable the hierarchy of the architecture. 
These example explore a two-layer network but the quality of the result suggest scalability towards more complicated hierarchies.

\section{Neuromorphic control of a five-link robotic snake}\label{sec:hierachical controller}

\subsection{Snake mechanics, muscle actuation, and neuromorphic controller basics}

We illustrate the flexibility and reliability of the proposed neuromorphic architecture by constructing a motion controller for a simplified, five-link snake robot. Our goal is to show that the functional network motifs presented in the previous section can be easily composed to build complex behavior across different scales. The illustrative purpose of this section is to suggest that the proposed architecture enables the design of a simple `nervous system' capable of driving the robot's locomotion.

\begin{figure}[H]
    \centering
    \includegraphics[width=0.6\linewidth]{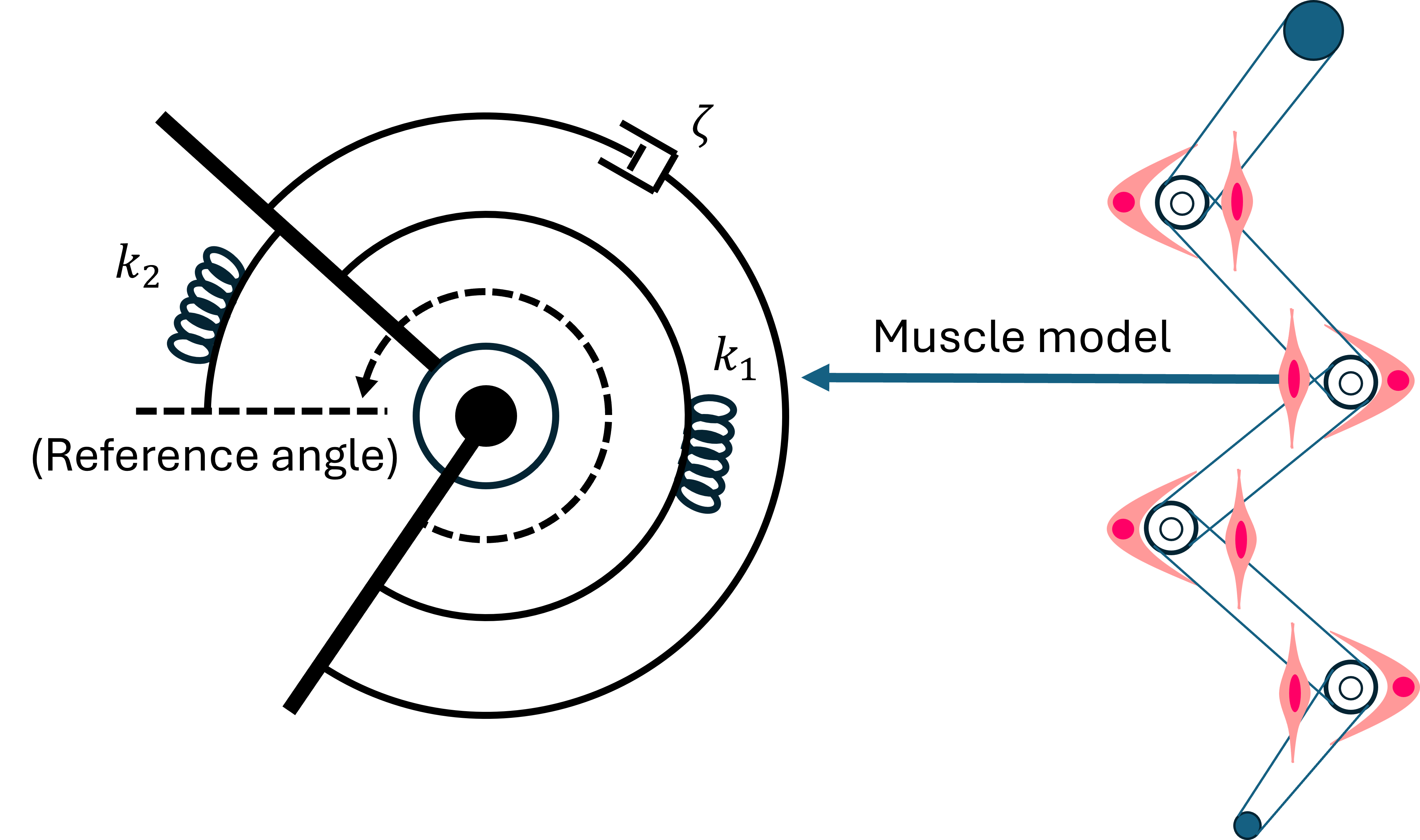}
    \caption{Plant: The five-link snake and muscle model}
    \label{fig:muscle_model}
\end{figure}

\underline{The plant}: the snake 'skeleton' consists of five rigid segments connected by four revolute joints. Each joint is actuated by a pair of `muscles'. Each muscle is modeled by (i) a torque spring with a variable reference point, (ii) a linear torque spring, and (iii) a linear damper \cite{mirbagheri_intrinsic_2000,feldman_once_1986}, as shown in Figure \ref{fig:muscle_model}. The contraction/relax of a muscle is modeled via the change in the reference trajectory of the first spring. In what follows, we regard the combination of snake mechanics and muscles as the \emph{plant}. 

\underline{The controller}: our neuromorphic architecture is used to design a hierarchical \emph{controller}. The controller outputs are the muscle reference trajectories $r$ on each side of each joint.
As summarized in Figure \ref{fig:intro}, the controller is organised into three layers: (i) Muscle Actuation Layer (local joint actuation) (ii) Coordination Layer (gait phasing and traveling waves) (iii) Supervisory Control Layer (task-level switching). All three layers use the same RWTA architecture. They differ only in topology and time-scale. Each layer is further organized into event, functional, and system levels. 

The controller architecture uses five different categories of synapses:
inhibitory ($I$),
slow inhibitory ($I_s$),
slow excitatory ($E$),
fast high-threshold excitatory ($E_f$),
fast low-threshold excitatory ($E_{fl}$),
and ultra-slow excitatory ($E_{us}$).
Their parameters are listed in Table~\ref{tab:syn_params}.
Fast ($\tau=0.5$), slow ($\tau=20$), and ultra-slow ($\tau=100$) synapses are coloured blue/red/black in the figures. Arrows denote excitation and hollow circles denote inhibition. 

\begin{table}[H]
    \centering
    \caption{Parameters of synapses.}
    \label{tab:syn_params}
    \begin{tabular}{lccc}
        \toprule
        Synapse type & $\tau$  & $\alpha$ & $\bar{V}$ \\
        \midrule
         $I/E_{fl}$     & 0.5  & 50 & -2.5 \\
         $E/I_s$       &  20  & 50 & -2.5 \\
         $E_f$       &  0.5 & 50 & 0 \\
         $E_{us}$     &  100 & 50 & -2.75 \\
        \bottomrule
    \end{tabular}
\end{table}

\subsection{Muscle Actuation Layer}\label{subsec:muscle_layer}

\noindent\underline{Goal.} 
Produce alternating left/right muscle activation at each joint, for local swinging.

\noindent\underline{Event level.}
At this level, the basic motor command is a \emph{muscle contraction event}.
We encode a contraction on one side of a joint as a \emph{burst}: a rebound burst drives one of the two sides ON for a given duration, while the other side is suppressed.
Because the burst duration is set by the ultra–slow current of the neuron, the duration is directly tunable.

\begin{figure}[H]
    \centering
    \includegraphics[width=0.8\linewidth]{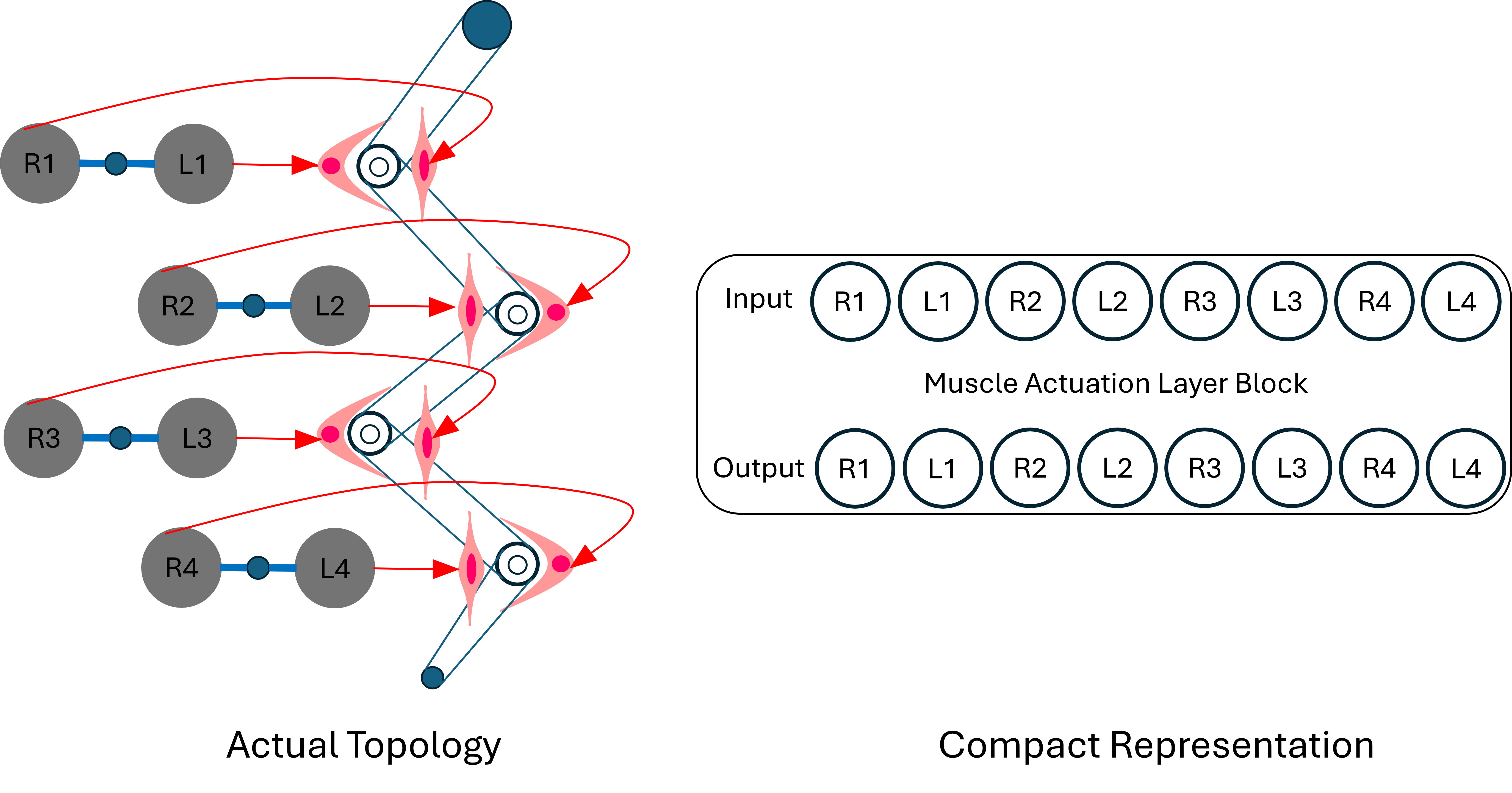}
    \caption{Muscle Actuation Layer.
    Each joint is controlled by a Half–Center Oscillator (HCO): two mutually inhibitory rebound neurons that alternate activity and drive the muscles via slow excitatory synapses $E$.
    Four HCOs (one per joint) form the Muscle Actuation Layer Block.}
    \label{fig:HCO_topo}
\end{figure}

\noindent\underline{Functional level.}
Each joint is driven by a \emph{half–center oscillator} (HCO), given by two rebound bursting neurons connected by fast reciprocal inhibition $I$. This produces left/right alternation of the muscle activation of each joint.
Each neuron in the HCO connects via a slow excitatory synapse $E$ to the corresponding muscle reference point $r$ (Figure \ref{fig:HCO_topo}).
The HCO block acts as a local muscle controller: one neuron excites the left muscle, and the other neuron excites the right muscle.

\begin{figure}[H]
    \centering
    \begin{subfigure}[t]{0.5\textwidth}
        \includegraphics[width=\textwidth]{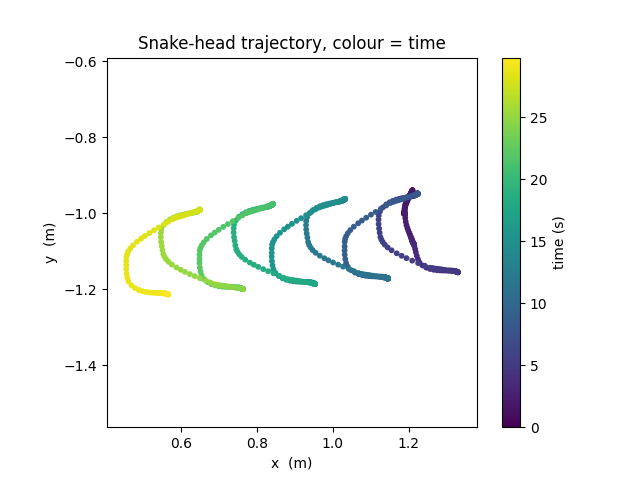}
        \caption{}
        \label{fig:HCO_only_trajectory}
    \end{subfigure}%
    \begin{subfigure}[t]{0.5\textwidth}
        \includegraphics[width=\textwidth]{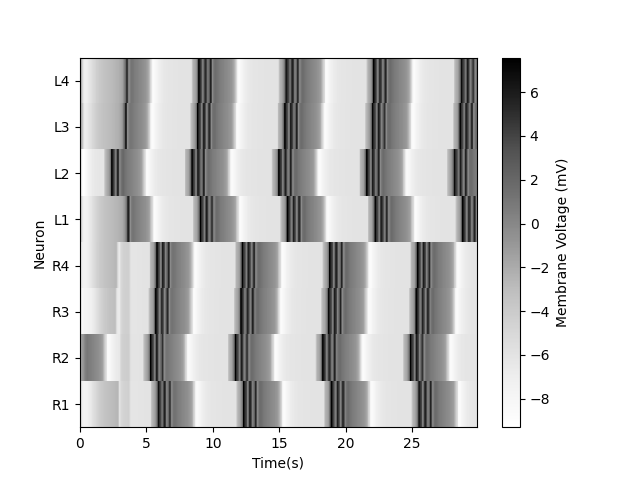}
        \caption{}
        \label{fig:HCO_only_raster}
    \end{subfigure}
    \vspace{-3pt}
    \begin{subfigure}[t]{0.5\textwidth}
        \includegraphics[width=\textwidth]{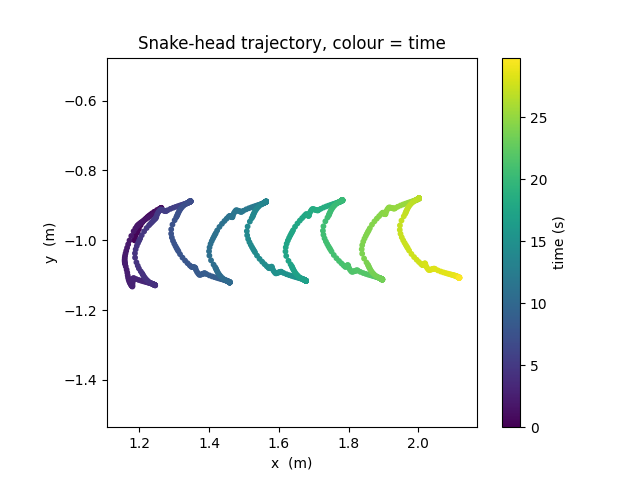}
        \caption{}
        \label{fig:HCO_only_trajectory_2}
    \end{subfigure}%
    \begin{subfigure}[t]{0.5\textwidth}
        \includegraphics[width=\textwidth]{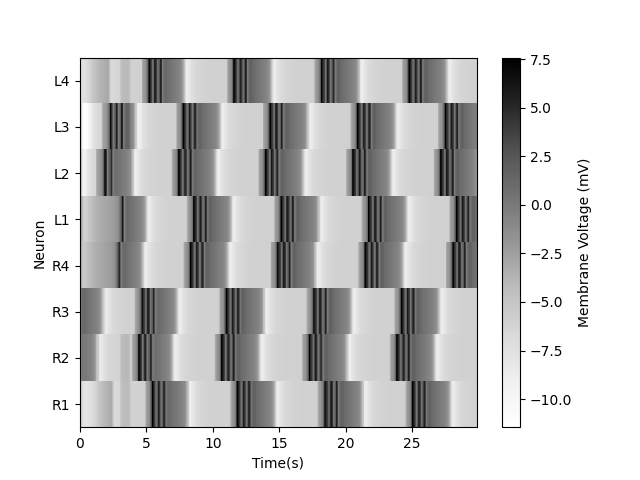}
        \caption{}
        \label{fig:HCO_only_raster_2}
    \end{subfigure}
    \caption{
    Muscle Actuation Layer in isolation.
    (a,c) Head trajectories of the snake under two different initial HCO phases.
    (b,d) Corresponding membrane potentials (raster-style) for the four uncoupled HCOs.
    The plant already moves, but the individual patterns still need to be orchestrated.}
\end{figure}

\noindent\underline{System level.}
We have four HCOs, one per joint. Together with the muscles, they form  \textbf{Muscle Actuation Layer Block}, represented in Figure
\ref{fig:HCO_topo}.
The coordination of these four
HCO, that is, the relative phase of the HCOs, is essential for locomotion. Different `initial conditions' produce distinct trajectories. This is shown in Figures \ref{fig:HCO_only_trajectory} and \ref{fig:HCO_only_trajectory_2}
show two trajectories generated by two different `initial conditions'. The corresponding membrane traces are reported in Figures \ref{fig:HCO_only_raster} and \ref{fig:HCO_only_raster_2},
repspectively. 
The coordination among the four HCOs can be regulated. This is the role of the next layer.

\subsection{Coordination Layer}\label{subsec:pattern_layer}

\noindent\underline{Goal.} Impose a forward/backward travelling wave across joints by (i) generating ordered spikes and (ii) converting them into HCO entrainment events.

\noindent\underline{Event level.}
At this level, we want to regulate the sequence of HCO activations. That is, beyond left/right activation, we want to regulate 'which joint should fire next'.
We represent this as a traveling wave of spikes across joints.
Each spike is an event that commands one joint to contract on one side.
We also use a single burst event to tune the overall cycle period.

\noindent\underline{Functional level.}
The Coordination Layer is based on two main  blocks:
\begin{description}
    \item[Forward/Backward Generation Block] 
    (Figure \ref{fig:pattern_generation}).
    We build two 5–node RWTA rings:
    a \emph{forward ring} and a \emph{backward ring}.
    Each ring has four spiking neurons (one per joint) and one bursting neuron (which sets the dwell time and thus the gait frequency).
    All ten neurons (both rings) share a fast all-to-all inhibitory hub. This guarantees that only one neuron across \emph{both} rings can be active at any given time (single-winner constraint).
    The forward ring produces a spike sequence in joint order 1$\to$2$\to$3$\to$4; the backward ring produces 4$\to$3$\to$2$\to$1.
    
\begin{figure}[H]
    \centering
    \includegraphics[width=0.8\linewidth]{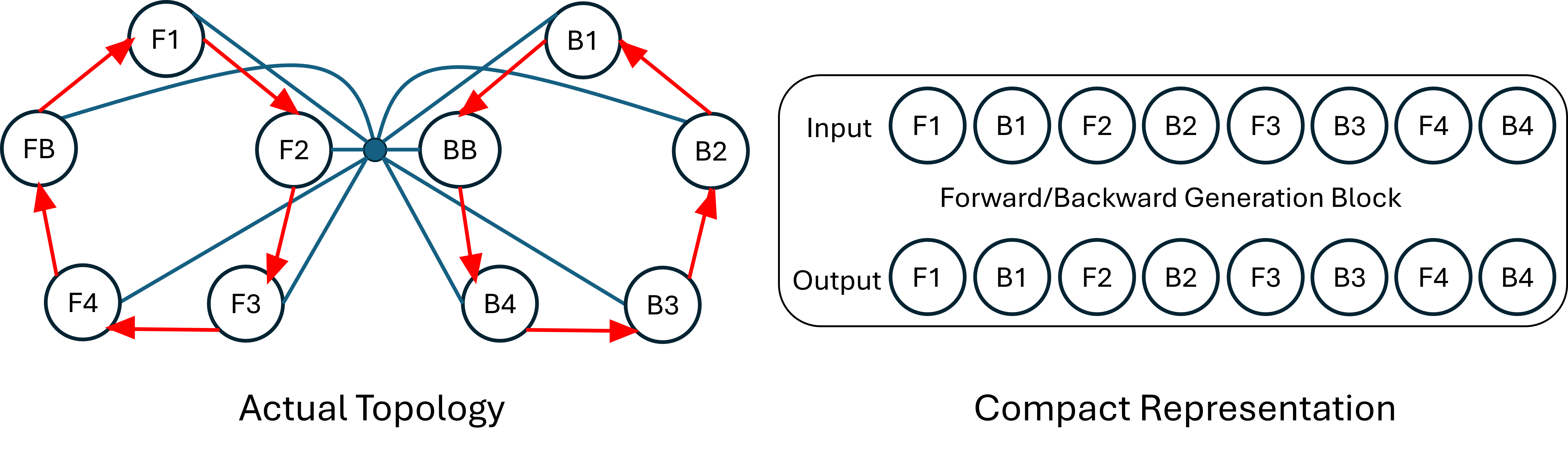}
    \caption{Coordination Layer: Forward/Backward Generation Block (two RWTA 5–rings with shared inhibition) and the Interface Module Block (RF/LF chains plus AND-like selectors). The active ring sets the order of joints; the interface selects which side of each joint fires.}
    \label{fig:pattern_generation}
\end{figure}

    \item[Interface Module Block] 
    (Figure \ref{fig:interface}).
    The spike events produced by the forward/backward generation block need to establish a consistent entrainment of the HCOs of the Muscle Actuation Layer. For each joint $k$, this is achieved by the interface module by the generation of two `entrainment events’: the spiking neuron RF$k$ inhibits the left HCO neuron L$k$ and excites the right HCO neuron R$k$. The effect is to pull the joint $k$ to the right. Similarly, the spiking neuron LF$k$ inhibits R$k$ and excites L$k$, pulling the joint to the left.
    
    The Interface Module decides to activate either the RF chain (RF1–RF4) and the LF chain (LF1–LF4). The activation is then propagated along the chain through the action of slow excitatory synapses.
    The activation decision is based on (i) the spike of the active ring of the forward/backward generation block, and on (ii) the associated HCO in the muscle actuation layer. For instance. The decision is implemented through four additional neurons, RFF, LFF, RFB, and LFB, acting as AND gates. The overall behavior of the network depends on the particular interconnection between modules, outlined in system level part.

\end{description}

\begin{figure}[H]
    \centering
    \includegraphics[width=0.6\textwidth]{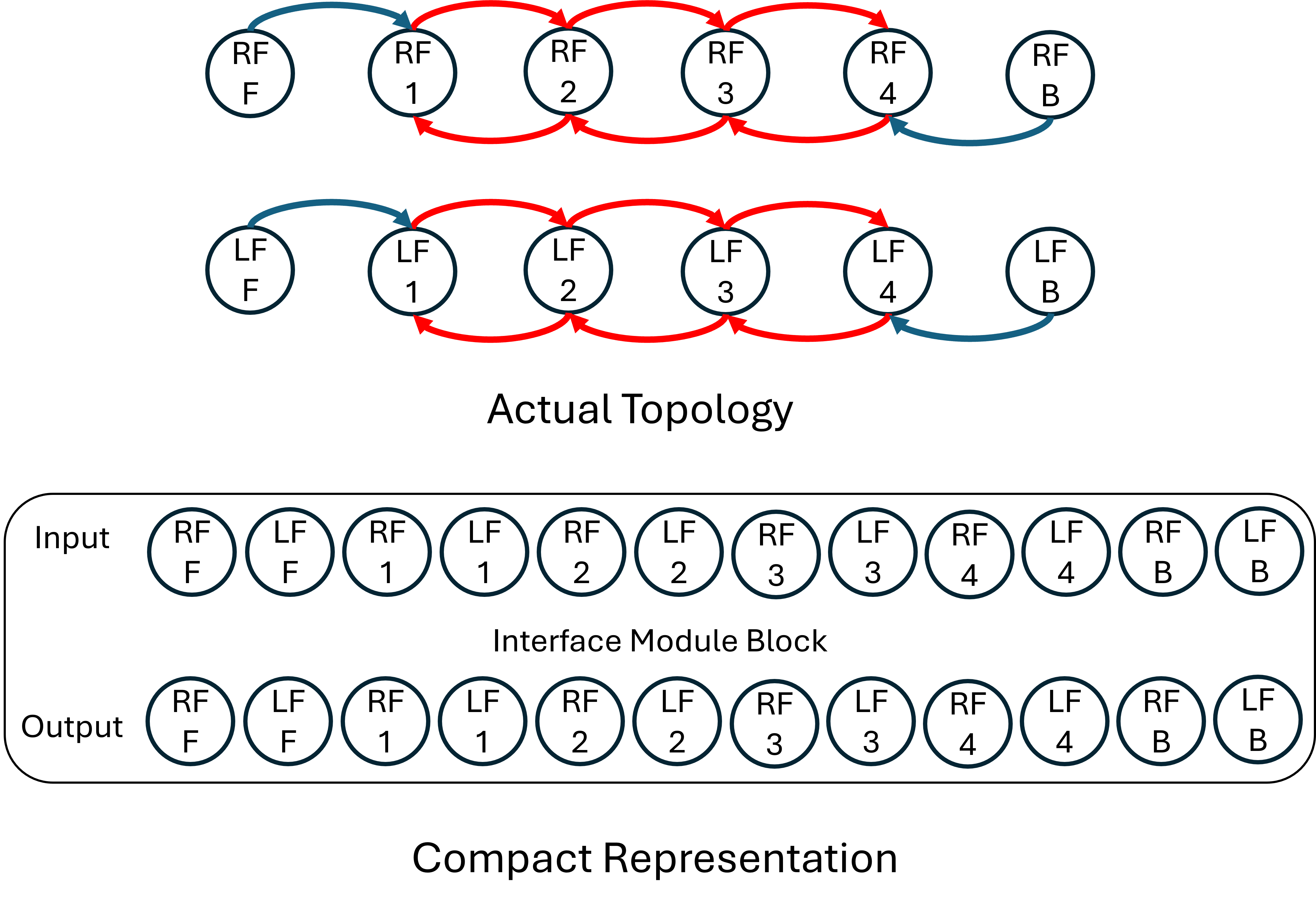}
    \caption{Interface Module Block.
    RF$k$ nodes bias joint $k$ towards ``right’’ (inhibit left HCO neuron, excite right).
    LF$k$ nodes bias joint $k$ towards ``left’’.
    Slow recurrent excitation within each chain (RF1$\leftrightarrow$RF2$\leftrightarrow$RF3$\leftrightarrow$RF4, similarly LF) maintains the same entrainment type for the duration of a cycle once chosen by the first spike.}
    \label{fig:interface}
\end{figure}

\begin{figure}[H]
    \centering
    \includegraphics[width=0.7\textwidth]{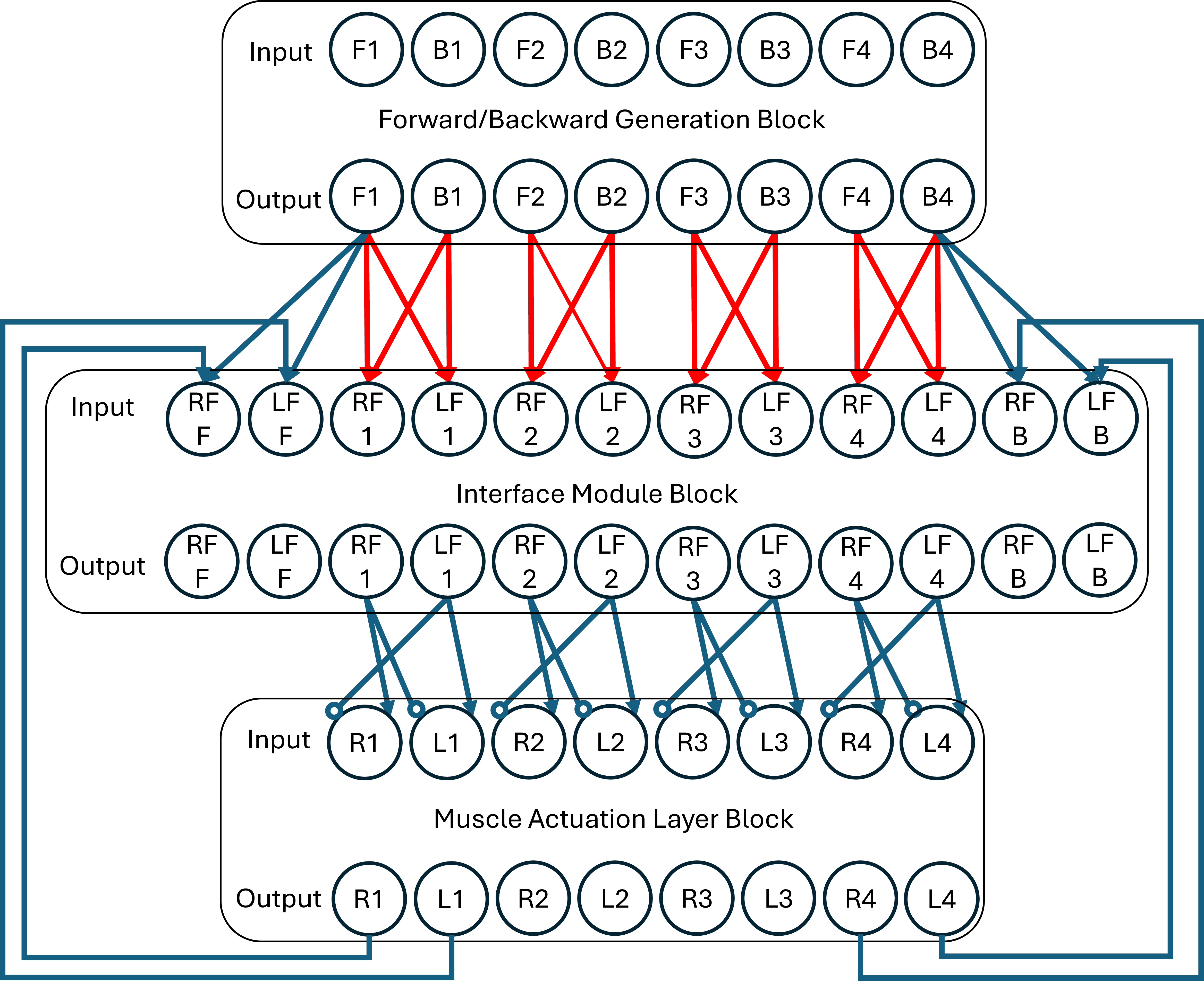}
    \caption{Full Coordination Layer interconnection.
    The Forward/Backward Generation Block provides ordered spikes.
    The Interface Module selects RF vs LF from the first spike and first HCO phase.
    The resulting entrainment events bias the HCOs in the Muscle Actuation Layer.}
    \label{fig:pattern_full}
\end{figure}

\noindent\underline{System level.}
The full coordination layer is obtained by interconnecting
(i) the forward/backward generation block,
(ii) the interface module block, and
(iii) the muscle actuation layer block,
as shown in Figure \ref{fig:pattern_full}.
The forward/backward generation block provides an ordered spike sequence (forward or backward wave), the interface block converts that sequence into joint-specific left/right entrainment events,
and these entrainment events directly bias each HCO of the muscle actuation block, so that all four joints follow a coherent pattern.
This conceptual picture is well represented by the top-to-bottom (excitatory and inhibitory) interconnections of Figure \ref{fig:pattern_full}.

The returning connections between the output of the muscle actuation layer block and the input of the interface module block are used by the interface module block to decide the activation of the RF chain (RF1–RF4) and the LF chain (LF1–LF4). For instance, in forward mode (when the forward ring of the forward/backward generation block is active), the spike of neuron F1 is combined (AND operation) to the spike of the HCO neurons R1/L1, related to joint 1. If R1 is more depolarized than L1 when F1 spikes, the RF chain is enabled (right–bias); otherwise the LF chain is enabled (left–bias). In a similar way, in backward mode (when the backward ring is active of the forward/backward generation block is active), the same decision involves neuron B4 and the HCO neurons R4/L4 of joint 4.

\begin{figure}[H]
    \centering
    \begin{subfigure}[t]{0.5\textwidth}
        \includegraphics[width=\textwidth]{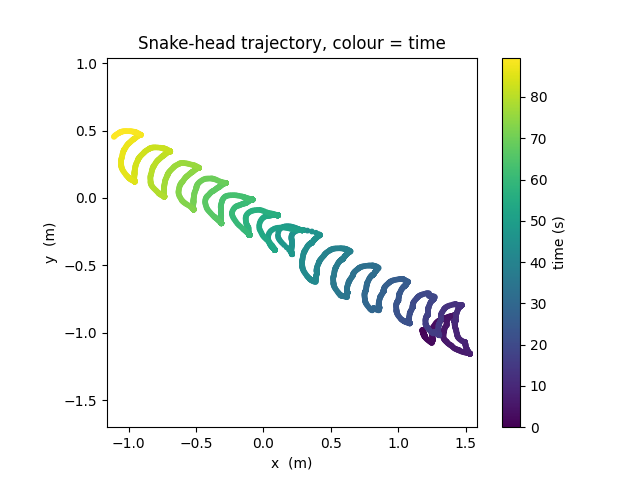}
        \caption{ }
        \label{fig:pattern_forward}
    \end{subfigure}%
    \begin{subfigure}[t]{0.5\textwidth}
        \includegraphics[width=\textwidth]{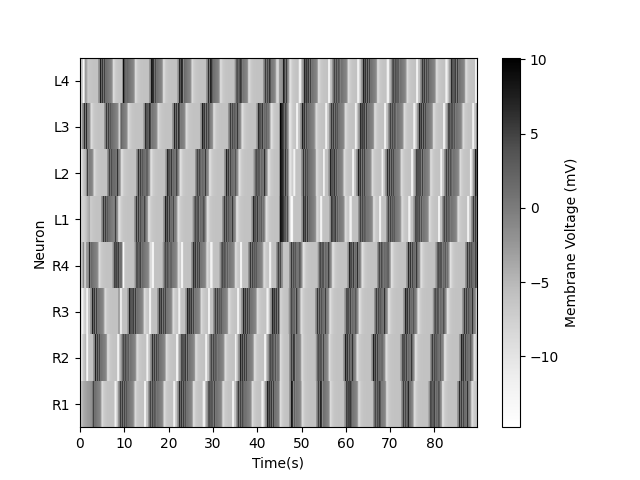}
        \caption{ }
        \label{fig:pattern_raster_HCO}
    \end{subfigure}
    \caption{Effect of the Coordination Layer.
    A traveling wave of joint activations produces coherent locomotion, and a single external event can re-phase the whole body. a) Snake head trajectory and b) Membrane potentials of the four HCOs with Coordination Layer enabled. At $t=45$\,s an external pulse resets the HCO phases, switching the selected entrainment chain.}
\end{figure}

Figure~\ref{fig:pattern_forward} illustrate the motion of the snake head when the coordination layer is active. The corresponding sequence of membrane potentials of the motor activation layer block neurons is
provided in Figure~\ref{fig:pattern_raster_HCO}.
At time $t=45$s an external reset is applied to the HCOs: R1-R4 are inhibited and L1-L4 activated.
This demonstrates how the sequence of muscle activations is on one hand globally coordinated and on the other hand can be re-synchronized by a single event.

\subsection{Supervisory Control Layer}\label{subsec:supervisory_layer}
\noindent\underline{Goal.} When bumping against an obstacle, execute \emph{right turn} or \emph{back retreat}. Alternate between the two in case of multiple consecutive obstacles.

\noindent\underline{Event level.}
At the task level, the system must choose between two distinct behaviors upon encountering an obstacle. Each behavior is encoded as a persistent ON event generated downstream of the spiking network. This is achieved by passing the spiking output of an HCO through an ultra-slow excitatory synapse $E_{us}$.
This synapse effectively functions as a low-pass filter, transforming the transient spikes into a sustained, step-like bias that signals the chosen behavior.

\begin{figure}[H]
    \centering
    \begin{subfigure}[b]{0.5\textwidth}
        \centering
        \includegraphics[width=1\linewidth]{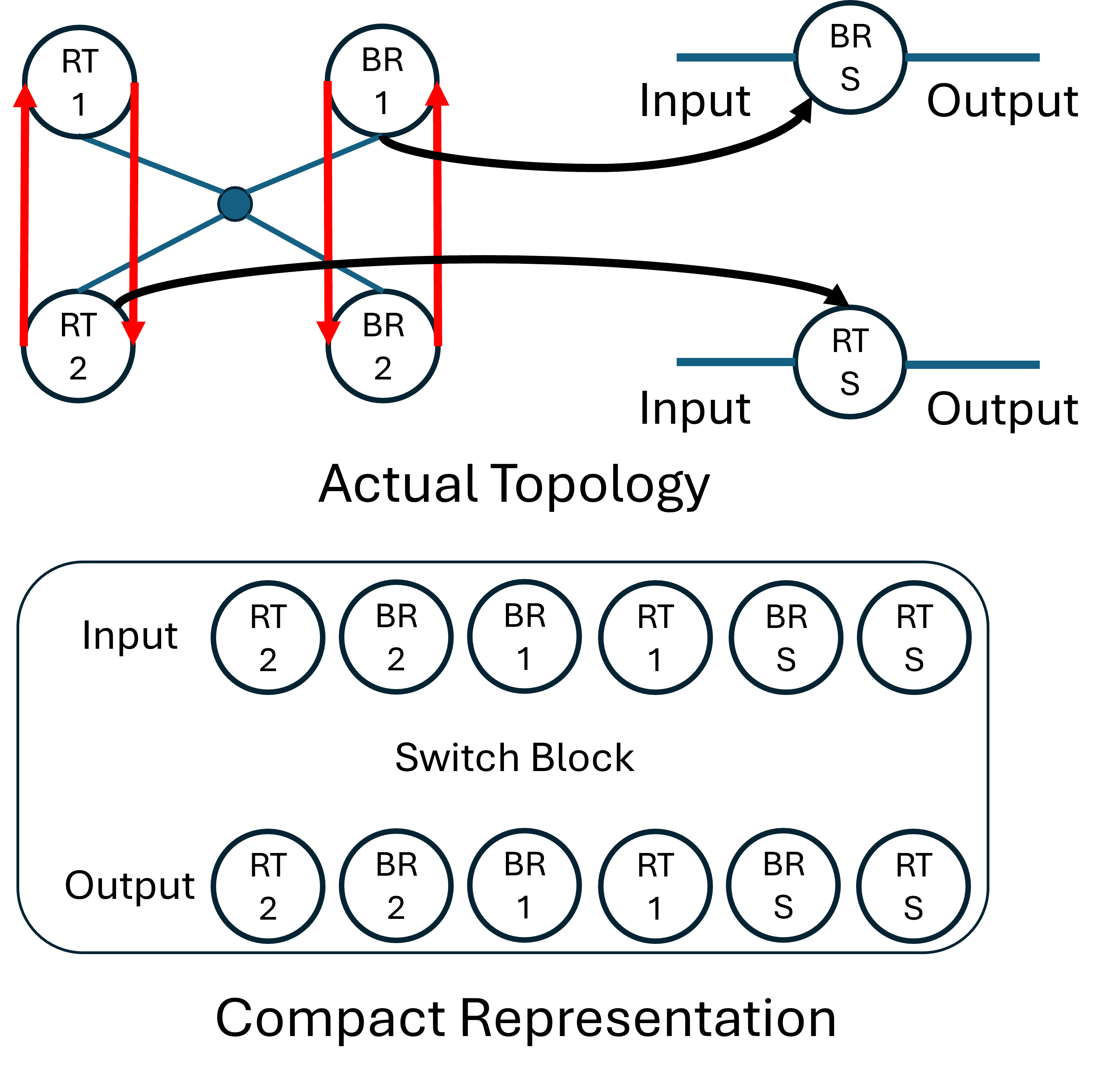}
        \caption{Two–state switch block.
        Ultra–slow excitation ($E_{us}$) from the currently active state biases either RTS (right–turn strategy armed) or BRS (back–retreat strategy armed).}
        \label{fig:switch_logic_topo}
    \end{subfigure}
    \hfill
    \begin{subfigure}[b]{0.45\textwidth}
        \centering
        \includegraphics[width=1\linewidth]{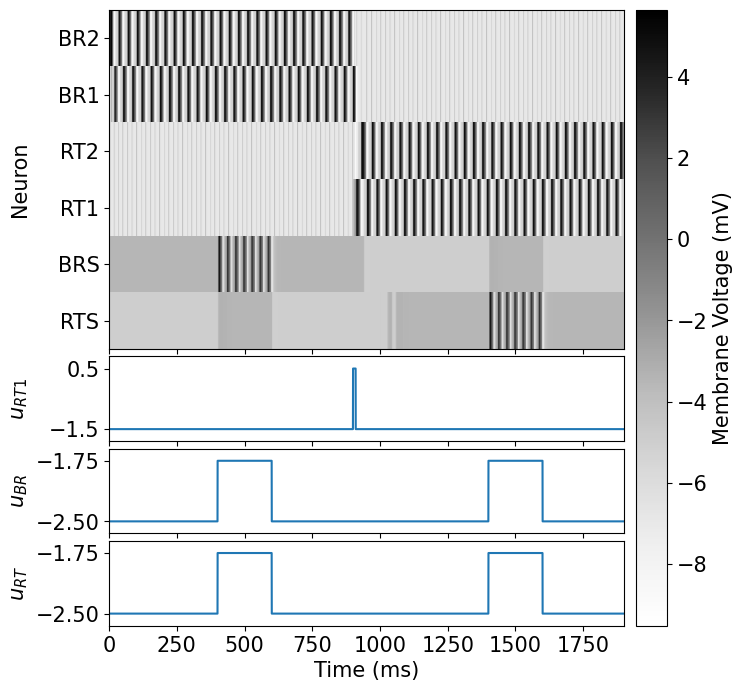}
        \caption{Switch timing.
        The same sensor input is applied twice, but between them a spike toggles the internal state, so different strategies are expressed.}
        \label{fig:switch_logic_function}
    \end{subfigure}
    \caption{Supervisory Control Layer: the switch block maintains which avoidance strategy is ``armed’’ and gates RTS/BRS accordingly.}
    \label{fig:switchfunction}
\end{figure}

\noindent\underline{Functional level.}
The first component of the supervisory control layer is given by the \textbf{Switch block},
whose function is to store which behaviour (right turn vs back retreat) is currently active. As shown in Figure \ref{fig:switch_logic_topo},
the switch block combines a RWTA network with ultra–slow synapses $E_{us}$ acting on two new interface neurons, RTS (right-turn) and BRS (back-retreat). 

The network's behavior is illustrated in Figure \ref{fig:switch_logic_function}. Activation of either the RT1-RT2 or BR1-BR2 pathway results in an ultra-slow output that biases the corresponding interface neuron, RTS or BRS, respectively. This bias primes the neuron to respond. For example, as shown in Figure \ref{fig:switch_logic_function}, even with identical inputs, the resulting activation levels of the RTS and BRS neurons are differentiated based on the current state of the RWTA network.

\begin{figure}[H] 
\centering 
\includegraphics[width=0.8\linewidth]{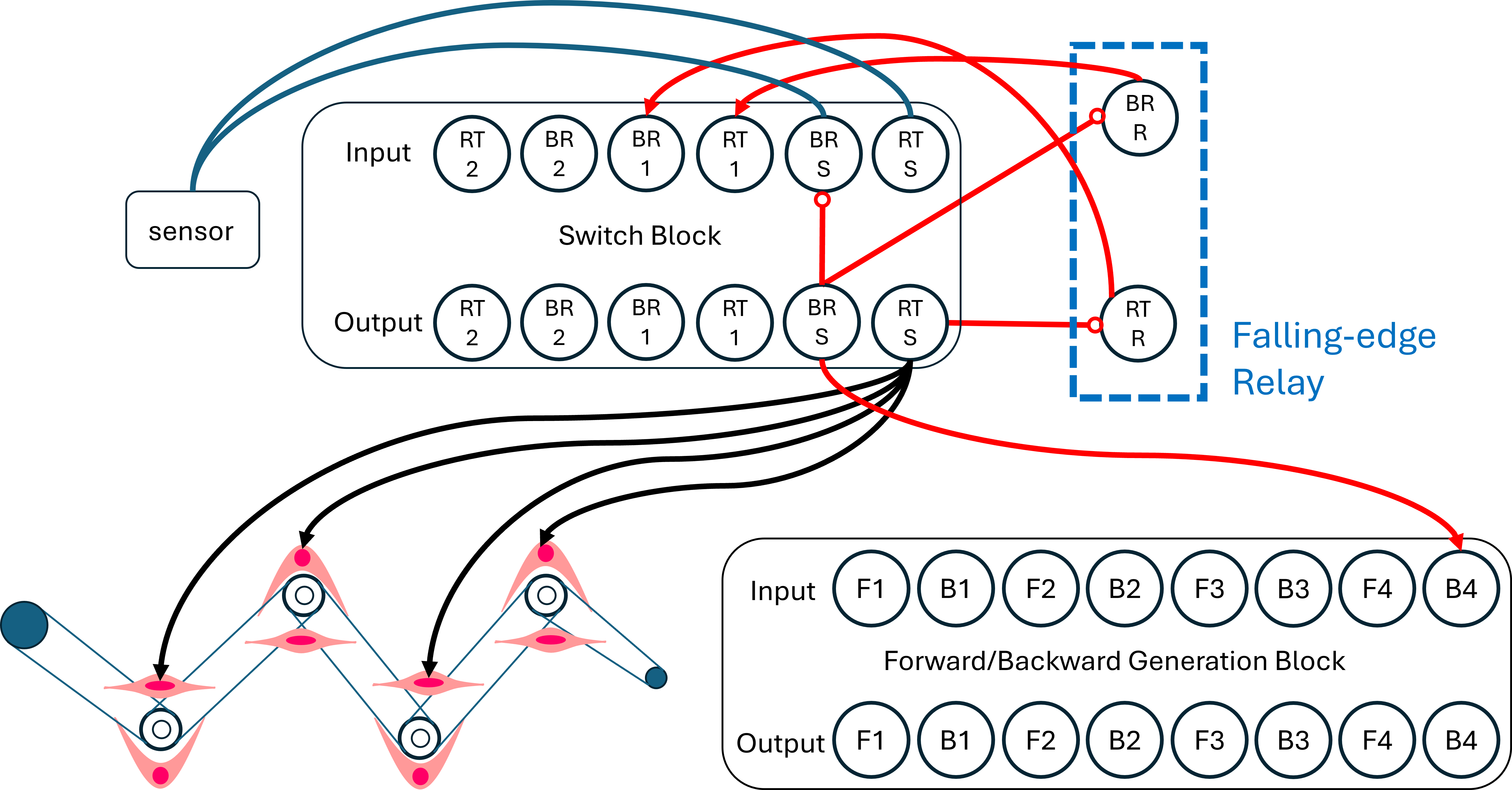} 
\caption{System-level view of the Supervisory Control Layer.
The tactile sensor excites RTS/BRS.
ultra-slow excitatory synapses from RTS directly biases the muscle on the right hand side for an right turn.
A slow excitatory synapses from BRS trigger the Coordination Layer (Backward ring) for a back retreat.
The ``falling-edge relay'' neurons toggle the two–state switch on contact release, forcing alternation between strategies.}
\label{fig:2-state} 
\end{figure}

The second component of the supervisory control layer is the \textbf{falling-edge relay} represented by the dashed box in Figure \ref{fig:2-state}. Its function is to  detect \emph{contact release}. When contact ends, each relay emits a rebound spike. This can be used to toggle the switch block, as outlined in the system level part below. The result is  the alternation: first obstacle → right turn, next obstacle → back retreat, next obstacle → right turn again, and so on.

\noindent\underline{System level.}
The supervisory control layer interact with 
both the coordination layer and the  muscle activation layer, as shown in Figure \ref{fig:2-state}.
The resulting behavior is described below.
When contact is detected,
\begin{itemize}[leftmargin=1.5em]
    \item[(i)] if neurons RT1 and RT2 are active,  the RTS neuron becomes active. This neuron is directly connected to the muscles of the
    right-hand side of the snake via ultra-slow excitatory synapses $E_{us}$.
    The result is an `in-place' right turn (Figure \ref{fig:2-state});
    \item[(ii)] if neurons BR1 and BR2
    are active, the BRS neuron also becomes active. It's slow excitatory synapse triggers neuron BR4 of the coordination layer, causing a gait change. An extra self inhibition is also added to \textbf{BRS} to avoid repeated triggering of \textbf{B4}.
\end{itemize}

\begin{figure}[H]
    \centering
    \includegraphics[width=0.5\linewidth]{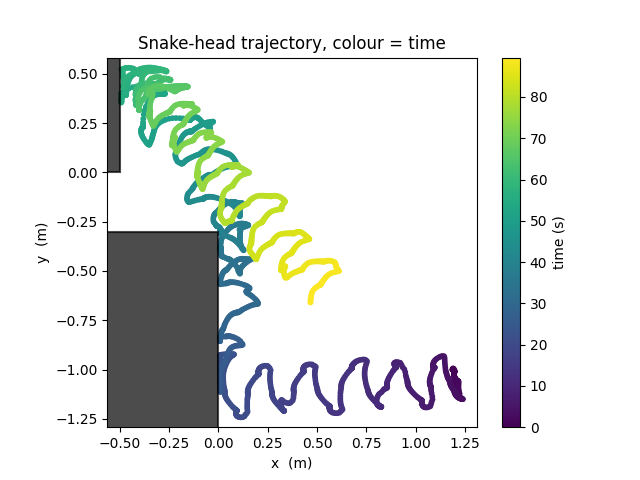}
    \caption{Behaviour under the Supervisory Control Layer.
    First obstacle contact: right turn. Contact ends, relay toggles internal state.
    Second obstacle contact: back retreat.}
    \label{fig:supervisory_right_back}
\end{figure}

When contact ends, the ``falling-edge relay'' spikes and toggles the switch, so that the next contact will trigger the other behaviour. This is illustrated in Figure \ref{fig:supervisory_right_back}. The first obstacle triggers a right turn; after release, the internal state flips; the second obstacle triggers a back retreat.

\begin{figure}[H]
    \centering
    \includegraphics[width=0.9\linewidth]{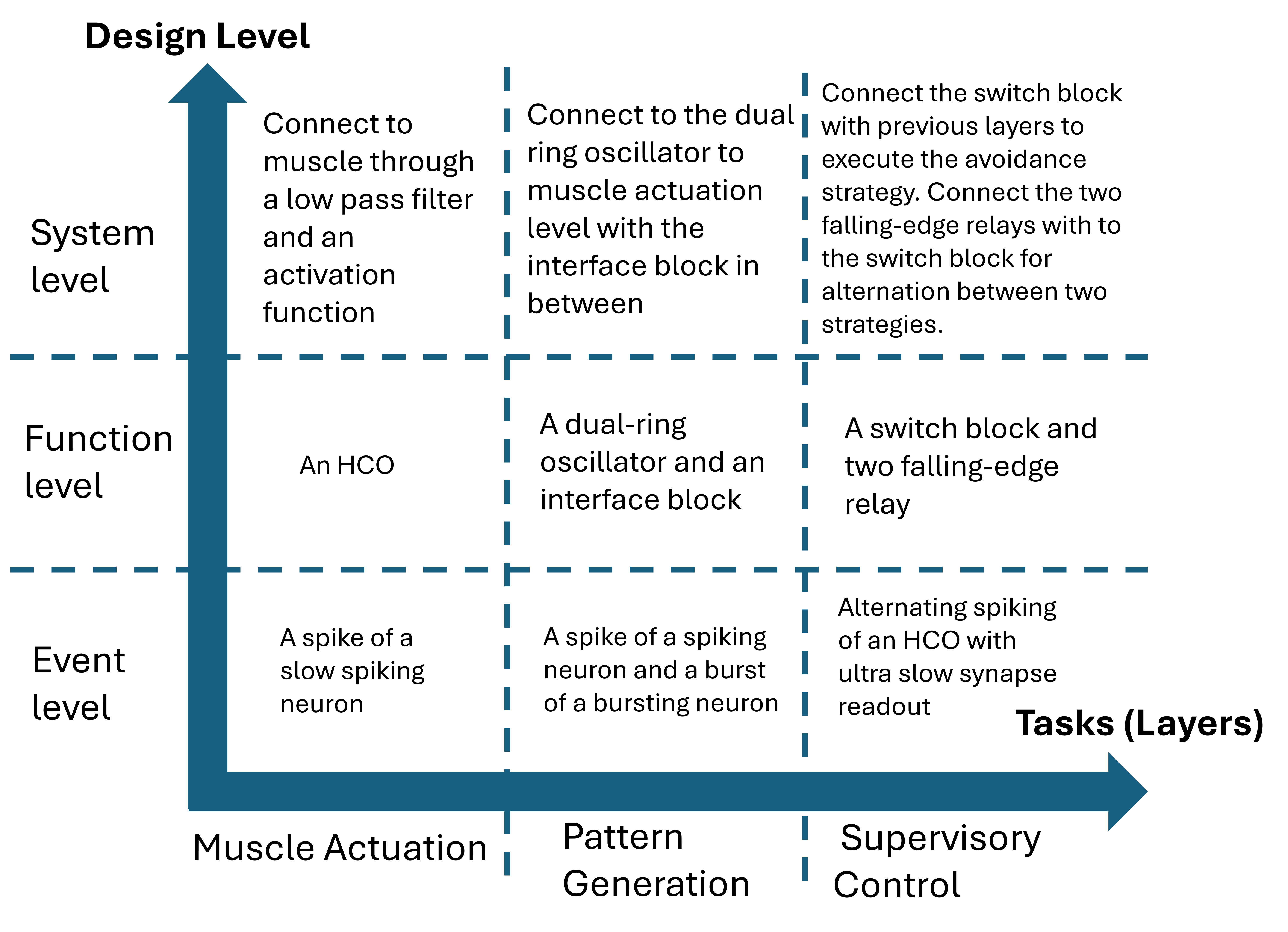}
    \caption{Summary of the controller architecture.
    The same RWTA design principles appear at three levels:
    Muscle Actuation Layer (local alternation),
    Coordination Layer (travelling wave and entrainment),
    Supervisory Control Layer (behavioural switching).
    All signals are voltage events (spikes or bursts); all interconnections are synaptic currents.}
    \label{fig:solution}
\end{figure}

\section{Discussion}\label{sec:discussion}
\subsection{Hierarchical event design}
Figure~\ref{fig:solution} can be regarded as a revision of the traditional design layout in Figure~\ref{fig:intro} in the event-based language of the present paper. A key feature of the rebound Winner-Take-All architecture is that all the cells in Figure~21 use the same modelling language and the same set of interconnection rules. Design, tuning, and adaptation are addressed at any level and any layer with the same concept of synaptically interconnected rebound neurons.

The central ingredients that make this architectural notion of scale concrete are (i) the separation of time-scales across the hierarchy and (ii) the separation of cellular excitability (event generation) from WTA-style competition (event orchestration). At the event level, all signals (sensory, internal, motor) are represented as spike or burst events generated by rebound-excitable neurons. Importantly, the semantics of an ``event'' evolve with scale: at the lowest level, events are spikes/bursts that encode changes in a physical variable or trigger a local actuation/entrainment mechanism; at higher levels, events can be interpreted as symbolic or supervisory decisions (e.g., ``enable forward wave'', ``enable right-turn strategy'').

This unification is expressed not only across different representations but also across different functions. The same rebound Winner-Take-All architecture can implement the rhythmic functions of pattern generation and the logical functions of an automaton. Higher level events consist of specific sequences of lower level events,  obeying the same modelling principles. This hierarchical event composition is what enables an end-to-end event-based representation that resembles the layered structure of conventional hiearchical control while remaining confined within one modeling language across scales.

Because all layers and levels share the same modeling language, they also share the same tuning properties: (i) changing neuron internal parameters to tune the shape and duration of an event (spike or burst); (ii) bias currents continuously tune the rebound time and thus rhythm frequency (demonstrated in Figure~9); (iii) sensory pulses, which apply immediate phase selection or hard resets by temporally biasing the current WTA competition. These three mechanisms are sufficient to modulate behaviours at every layer without introducing layer-specific control laws. As a result, adaptation is uniform: the same type of parameter change (bias or synaptic time constant/gain) has predictable, monotone effects from lower layer (muscle actuation) to higher layer (supervisory control).

Finally, it is worth emphasizing how this differs from existing architectures. Conventional neuromorphic locomotion tasks typically couple (i) a continuous CPG for rhythms, (ii) a discrete finite-state machine for mode logic, and (iii) separate low-level joint controllers, glued together with cross-domain interfaces \cite{lozano_control_2016,linares-barranco_towards_2022,zhang_online_2024}. This heterogeneity forces different design languages (ODEs for regulation, automata or associative memories for logic \cite{cotteret_vector_2024,cotteret_distributed_2025}), incurs interface fragility, and complicates tuning (parameters have layer-specific meanings). In contrast, the proposed framework avoids the need for ad-hoc glue logic between rate-based and event-based blocks, and avoids a change of modelling paradigm between decision making and trajectory planning. The end-to-end event-based representation makes the supervisory layer naturally interpretable as event-driven mode selection, which also invites connections to hybrid and variable-structure viewpoints (e.g., switching surfaces and mode logic), while still grounding the implementation in the same excitable physical primitives.

\color{black}
\subsection{A neuromorphic architecture...}

The central feature of the architecture proposed in this paper is that it is rooted in the physiology of {\it neuronal excitability}, and, singularly, {\it rebound} excitability. To the best of the authors knowledge, it is the first proposed event-based design architecture of this type, and it makes it eminently {\it neuromorphic}. Neuronal excitability is a core feature of any nervous system, but it has not been embraced by artificial intelligence technology. Our research suggests the potential of neuronal excitability for neuromorphic engineering:

\begin{enumerate}
    \item 
The design of an event-based behavior requires the design of {\it events} and the design of {\it sequences of events}. Neuronal excitability is the property that enables a complete decoupling between the two: shaping the event is achieved by regulating internal conductances, whereas shaping the sequence of events is achieved by regulating external conductances. The hierarchy of our architecture comes from the fact that events at coarser scales are made of sequences of events at finer scales. For this reason, the decoupling between {\it internal} and {\it external} repeats itself across scales, with an evolving semantics for the functionality of events: spike control regulates burst events, burst control regulates pattern events, pattern control regulates locomotion events, locomotion control regulates behavior, etc..

\item The unique property of excitability to regulate {\it across scales} has been advocated in neurophysiology \cite{marom_biophysical_2023}, in computational neuroscience \cite{axenie_antifragile_2025}, and in control \cite{sepulchre_control_2019}. Our architecture is consistent with those principles. At every scale, an event is modeled by a neuromorphic circuit with the same conductance architecture, enabling regulation at every scale from the same adaptive mechanisms \cite{burghi_system_2020,schmetterling_adaptive_2022}. At every scale, the topology of the network constrains the sequence of events from the same complementary roles of excitatory and inhibitory synaptic interconnections. Those principles seem aligned with those long advocated in biological cybernetics \cite{wilson_spikes_1999}.

\item Excitability seems also unique in making heterogeneity a design feature rather than a design nuisance: events are tuned independently from each other, de facto resulting in populations of heterogenous events at any scale. In turn, this very heterogeneity becomes a {\it tuning} feature for the sequence of events that will define the event at the next scale. This feature was illustrated at an elementary level by tuning the period of a ring oscillator by inserting a burster in a ring of spiking neurons. Again, this concrete hierarchical tuning procedure seems consistent with recent studies on the value of heterogeneity in tuning and learning \cite{habashy_adapting_2024}. 
\end{enumerate}
\subsection{... for scalable event-based control}

There is a long way from designing a scalable architecture to designing a scalable machine. The present paper should only be regarded as a first and conceptual step towards scalable event-based control. We envision however that the scalability of the control design will benefit from the scalability of the architecture. In particular, the following three distinct avenues provide encouraging research directions:
\begin{enumerate}
\item Modeling, analysis, and synthesis {\it at scale}: a hierarchical architecture calls for hierarchical modeling, analysis, and design tools. Shaping an event at a coarser scale must be abstracted from shaping the event at a finer scale. The article \cite{scheres_discrete_2025} is a preliminary step in that direction: a hierarchy of {\it discrete} event models is extracted from the same {\it continuous} neuromorphic system to enable tractability {\it across} scales from tractability {\it at} scale. 
\item Tuning and learning {\it at scale}: classical principles from adaptive control are sufficient to tune  at cellular scale \cite{burghi_system_2020,schmetterling_adaptive_2022}. Currently, those principles do not extend {\it across} scales. We envision however that the same principles could be implemented at {\it any} scale, pending a systematic methodology to model the neuromorphic system at different scales. Similarly, classical Hopfield learning paradigms seem applicable at one scale (for instance, to learn a particular sequence from a all-to-all inhibitory architecture), and we envision to apply those same learning principles at any scale. This particular research direction will be the topic of a forthcoming publication by the authors. 
\item Physical realisations {\it at scale}. The question of turning the conceptual architecture into a physical realisation is not addressed in this paper but it deserves particular attention. The proposed architecture is an interconnection of memristive elements. At each scale, the design of new events requires memristive elements with commensurate  temporal and spatial scales: spikes require fast conductances, bursts require slower conductances, sequences of bursts require even slower synapses, and so on. 

The same hierarchy is found in the current research on memristive materials. Each physical domain comes with its own physical scales. It is a challenge to cover many scales in one specific domain, but much promise arises from interconnecting memristive devices across different physical domains (electronics, organics, pneumatics,...), and meristive designs are investigated in electronics, pneumatics, organics. Encouraging steps have been taken in implementing rebound bursting neurons on neuromorphic chips \cite{khan_brain_2025,wang_neuromorphic_2026}. 
 Research labs are currently exploring similar realisations in the pneumatic and organic domain. Our hierarchical architecture is well suited to exploring multi-scale implementations by exploiting multi-scale physical domains. 

 Besides physical realisations of neuromorphic architectures, more and more attention is also devoted to digital realisations of neuromorphic architectures. A digital implementation of the proposed neuromorphic architecture on dedicated chips such as \cite{frenkel_65-nm_2019} is a most promising avenue in the near future.
\end{enumerate}

\color{black}

\section{Conclusion}\label{sec:conclusion}

\noindent
This paper has introduced \emph{rebound winner--takes--all}  networks as a unifying primitive for decision making and control, The neuromorphic architecture was illustrated on the design of a snake motion task. By designing the topology, we created a range of rebound winner--takes--all modules that deliver functions at different layers: state network, pattern generation and muscle actuation without the need for extra ingredients. And these three layers combine to form a complete controller for the snake motion task.

\vspace{0.8em}
\noindent
The framework opens a promising route toward large-scale neuromorphic event-based architectures. Because all computations are carried through physical voltages and currents variables, sensory signals can be injected \emph{directly} into the network and motor commands can be read out \emph{directly} from synaptic waveforms, eliminating the need for "recording and processing" and hence make the computation truly embodied and event-based.  

From a controller point of view, the tunability of the network is not affected by the complexity of the topology---It is fully decentralized and distributed. This makes it appealing for event-based control as adapting the action time to the event time is completely decoupled from the orchestration task of the high level design.

\vspace{0.8em}
\noindent
Rebound Winner-Take-All networks sit at the \emph{intersection} of continuous (physical) and discrete (algorithmic) paradigms.  Winner-Take-All competition, long used as an analogue current-mode maximum selector
, endows the network with logical decision capabilities, while continuous rebound dynamics enable tuneable rhythms unavailable to purely digital logic.  


\vspace{0.8em}
\noindent

There are many questions that require research beyond the material of this paper. Primarily and foremost, we wish to develop learning and adaptation algorithms that can take advantage of the proposed neuromorphic architecture. The event-based nature of the neural network suggests the potential of Hebbian rules for learning and adaptation.  Another key question is the hardware realisation of the "nervous system" architecture, and the respective advantages of digital and analog devices all the way from sensing to actuation. Those questions will be addressed in a future publication.

\section{Appendix}
\subsection{Parameters for simulations and implementation detail for snake controller}\label{sec:params_mats_full} All spiking neurons used in this paper has the following parameters: $\tau_s=20$, $\alpha_f^-=\alpha_s^+=2$, $R=2$, ${\bar V}=-1.5$. For bursting neuron in Figure \ref{fig:bursting_neuron_2}, we use: $\tau_s=5$, $\tau_{us}=100, \alpha_f^-=\alpha_s^+=2$, $\alpha_s^-=1.5$, $R=0.5,{\bar V}_f=3$, ${\bar V}_s=-1.5$. We use $\alpha_{us}^+=1.5$ for trajectory A and $\alpha_{us}^+=1.7$ for trajectory B. 

In Figure \ref{fig:pattern_generation}, for HCO and the case of single inhibitory synapse the parameters are chosen to be  \(\tau = 0.5\), \({\bar V}=-1\) and \(\alpha=50\) for inhibitory synapses with weight $ -2$. All neurons are given a constant external input of -1.5. In Figure \ref{fig:pattern_generation}, for other plots, all inhibitory synapse used the same parameters: \(\tau = 0.5\), \({\bar V}=-2.5\) and \(\alpha=50\) with weight $ -2$. All excitatory synapses used  \(\tau = 20\), \({\bar V}=-2.5\) and \(\alpha=50\). All neurons are given a constant external input of -1.5.

In Figure \ref{fig:hierarchy}, all fast inhibitory synapse (blue lines) used the same parameters: \(\tau = 0.5\), \({\bar V}=-2.5\) and \(\alpha=50\) with weight $ -2$. All slow inhibitory synapses (purple lines) used  \(\tau = 10\), \({\bar V}=-3\) and \(\alpha=50\) and synaptic weight -2.5. All slow excitatory synapses (red arrow) used  \(\tau = 20\), \({\bar V}=-2.5\) and \(\alpha=50\) and synaptic weight 0.75. All ultra-slow inhibitory and excitatory synapse (black lines and arrows) used the same parameters: \(\tau = 200\), \({\bar V}=-4\) and \(\alpha=50\) with synaptic weight -2 for inhibition and synaptic weight 0.5 for excitation. All neurons are given a constant external input of -1.5.


 All the neurons and synapses are slowdown-ed by 3/100 times in the snake simulation. 
\subsubsection*{Muscle Actuation Layer}
In this layer, the bursting neurons have the following parameters: $\tau_s=5$, $\tau_{us}=100, \alpha_f^-=\alpha_s^+=2$, $\alpha_s^-=\alpha_{us}^+=1.5$, $R=0.5,{\bar V}_f=3$, ${\bar V}_s=-1.5$. 
Unless specified, the constant input for bursting neurons in this layer is  $u=-1$. Each HCO use the $I$ synapse as inhibition synapse with synaptic weight of -2. The input of the muscle is directly connected to a synapse with $\tau=20, \alpha=16,{\bar V}=-1.875$ and gain 1.5. 

\subsubsection*{Coordination Layer} 
In this layer, the bursting neurons have the following parameters: $\tau_s=5$, $\tau_{us}=100, \alpha_f^-=\alpha_s^+=2$, $\alpha_s^-=1.1$, $\alpha_{us}^+=1.7$, $R=0.5$, ${\bar V}_f=3$, ${\bar V}_s=-1.5$. 
For neurons in the ring oscillators (both spiking and bursting neuron), we use constant input $-0.5$. Fast all-to-all inhibition across the two rings are $I$ synapses with synaptic weight of -2.5. The slow excitatory connections in the two rings are $E$ synapses with synaptic weight of 0.75

Neurons in the interface module are given constant input of $-2$. The slow excitatory synapse (denote as red arrow in Figure \ref{fig:interface}) are $E$ synapse with synaptic weight 1.5. The fast excitatory synapse (denote as blue arrow in Figure \ref{fig:interface}) are $E_fl$ synapse with synaptic weight 1.5. 

Figure \ref{fig:pattern_full} shows the connections between the ring oscillators, the interface module and the muscle actuation layer. The red arrows are slow excitatory connections $E$ with synaptic weight 1. The blue arrows from F1, R1, L1, B4, R4, L4 to RFF, LFF, RFB, LFB are $E_f$ synapses. Specifically, the connections from F1, B4 to RFF, LFF, RFB, LFB have synaptic weight 1. The connections from R1, L1, R4, L4 to RFF, LFF, RFB, LFB have synaptic weight of 0.6. The blue arrows from RF1, LF1, RF2, LF2, RF3, LF3, RF4, LF4, are $E_{fl}$ synapses with synaptic weight 1.5. The blue circle head arrows from RF1, LF1, RF2, LF2, RF3, LF3, RF4, LF4, are $I$ synapses with synaptic weight -3.

\subsubsection*{Supervisory Control Layer}
RT1, RT2, BR1, BR2, BRR, RTR are given constant input of -0.5. BRS and RTS are given constant input of -2. Neurons in the 2-state machine (RT1, RT2, BR1, BR2) are all-to-all inhibited with $I$ synapses with synaptic weight of -2.5. The slow excitatory connection between them are denoted by red arrows in Figure \ref{fig:2-state} with synaptic weight of 0.75. The sensor input to BRS and RTS is either ON (1) or off (0). The slow excitatory connection ($E$) between BRS and B4 has synaptic weight of 2. The slow self inhibition ($I_s$) in BRS has synaptic weight -1.5. The slow inhibition ($I_s$) from BRS, RTS to BRR, RTR have synaptic weight of -1. The slow excitatory connection ($E$) from BRR, RTR to RT1, BR1 have synaptic weight of 2. The ultra-slow activation ($E_{us}$) from RT2, BR2 to BRS, RTS have synatic weight of 1. The ultra-slow connection between RTS and the muscles have $\tau=100, \alpha=16,{\bar V}=2.625$ and synaptic weight of 0.5.

For Figure \ref{fig:supervisory_right_back}, the $\alpha_{us}^+$ of the bursting neurons in the muscle actuation layer is adjusted to 2.75.

\printbibliography[title={References}]

\end{document}